\documentclass[10pt,twocolumn,letterpaper]{article}

\usepackage[pagenumbers]{cvpr} 

\usepackage[dvipsnames]{xcolor}

\usepackage[dvipsnames]{xcolor}
\usepackage{colortbl}
\usepackage{url}
\definecolor{cvprblue}{rgb}{0.21,0.49,0.74}
\usepackage[pagebackref,breaklinks,colorlinks,citecolor=cvprblue]{hyperref}
\usepackage{amsmath}
\usepackage{multirow}
\usepackage{float}
\usepackage{cuted}
\usepackage{bm}
\usepackage{comment}
\usepackage{caption}
\usepackage{balance}
\usepackage{pifont}
\usepackage[normalem]{ulem}
\usepackage{arydshln}
\usepackage{lipsum}
\usepackage{pifont}
\usepackage{stackengine}
\usepackage{fontawesome}
\usepackage[toc,page]{appendix}
\usepackage[accsupp]{axessibility}  

\newcommand{\methodname}{\methodCOLOR{SIFU}\xspace}
\newcommand{\mn}{\methodname}

\definecolor{bittersweet}{rgb}{1.0, 0.44, 0.37}
\newcommand{\methodCOLOR}[1]{{\color{black} #1}}

\definecolor{bestcolor}{rgb}{1, 0.7, 0.5} 
\definecolor{secondbestcolor}{rgb}{1, 0.9, 0.7} 
\newcommand{\bone}{\cellcolor{bestcolor}}
\newcommand{\btwo}{\cellcolor{secondbestcolor}}

\definecolor{todocolor}{RGB}{255,0,00}

\definecolor{DeltaColor}{rgb}{0.039,0.73,0.71}
\definecolor{SigmaColor}{rgb}{0.98,0.45,0.0}
\definecolor{TODOColor}{rgb}{0.98,0.0,0.0}
\definecolor{AlphaColor}{rgb}{0,0,0.8}
\definecolor{BetaColor}{rgb}{0.8,0,0.8}
\definecolor{GammaColor}{rgb}{0.514,0.34,0.224}
\definecolor{EpsilonColor}{rgb}{0.353,0.725,0.906}
\definecolor{citecolor}{HTML}{0071bc}

\newlength\savewidth\newcommand\shline{\noalign{\global\savewidth\arrayrulewidth
  \global\arrayrulewidth 1pt}\hline\noalign{\global\arrayrulewidth\savewidth}}

\DeclareTextFontCommand{\specific}{\small\fontfamily{qcr}\selectfont}

\DeclareMathOperator*{\argmin}{arg\,min}
\newcommand{\colorRef}[1]{\textcolor{red}{#1}} 
\usepackage[capitalize]{cleveref}
\crefname{figure}{\colorRef{Fig.}}{\colorRef{Figs.}}
\Crefname{figure}{\colorRef{Figure}}{\colorRef{Figures}}
\crefname{section}{\colorRef{Sec.}}{\colorRef{Secs.}}
\Crefname{section}{\colorRef{Section}}{\colorRef{Sections}}
\Crefname{table}{\colorRef{Table}}{\colorRef{Tables}}
\crefname{table}{\colorRef{Tab.}}{\colorRef{Tabs.}}

\newcommand{\ourrefcolor}[1]{\textcolor{red}{#1}} 

\newcommand{\tab}[1]{\mbox{\ourrefcolor{{Tab.}}~\ref{#1}}}

\newcommand{\sect}[1]{\mbox{\ourrefcolor{{Sec.}}~\ref{#1}}}

\newcommand{\smplx}{\mbox{SMPL-X}\xspace}

\newcommand{\thtwo}{\mbox{THuman2.0}\xspace}

\newcommand{\cmark}{\ding{51}}%
\newcommand{\xmark}{\color{red}{\ding{55}}}%

\newcommand{\pifu}{\mbox{PIFu}\xspace}
\newcommand{\pamir}{\mbox{PaMIR}\xspace}
\newcommand{\phorhum}{\mbox{PHORHUM}\xspace}
\newcommand{\pifuhd}{\mbox{PIFuHD}\xspace}
\newcommand{\econ}{\mbox{ECON}\xspace}
\newcommand{\icon}{\mbox{ICON}\xspace}
\newcommand{\dif}{\mbox{D-IF}\xspace}
\newcommand{\gta}{\mbox{GTA}\xspace}
\DeclareSymbolFont{matha}{OML}{txmi}{m}{it}
\DeclareMathSymbol{\varv}{\mathord}{matha}{118}

\def\paperID{****} 
\def\confName{CVPR}
\def\confYear{2024}

\title{SIFU: Side-view Conditioned Implicit Function for Real-world Usable \\ Clothed Human Reconstruction}

\author{Zechuan Zhang \hspace{1.5em}  Zongxin Yang\footnotemark[2] \hspace{1.5em}  Yi Yang \\
ReLER, CCAI, Zhejiang University\\
{\tt\small \{zechuan, yangzongxin, yangyics\}@zju.edu.cn}
}

\begin{document}
\newcommand{\teaserCaption}{
With just a single image, \mn is capable of reconstructing a high-quality 3D clothed human model, making it well-suited for practical applications such as 3D printing and scene creation. At the heart of \mn is a novel \textbf{Side-view Conditioned Implicit Function}, which is key to enhancing feature extraction and geometric precision. Furthermore, \mn introduces a \textbf{3D Consistent Texture Refinement} process, greatly improving texture quality and facilitating texture editing with the help of text-to-image diffusion models. Notably proficient in dealing with complex poses and loose clothing, \mn stands out as an ideal solution for real-world applications.
}

\twocolumn[{
    \renewcommand\twocolumn[1][]{#1}
    \maketitle
    \centering
    \begin{minipage}{1.00\textwidth}
        \centering 
        \includegraphics[width=\linewidth]{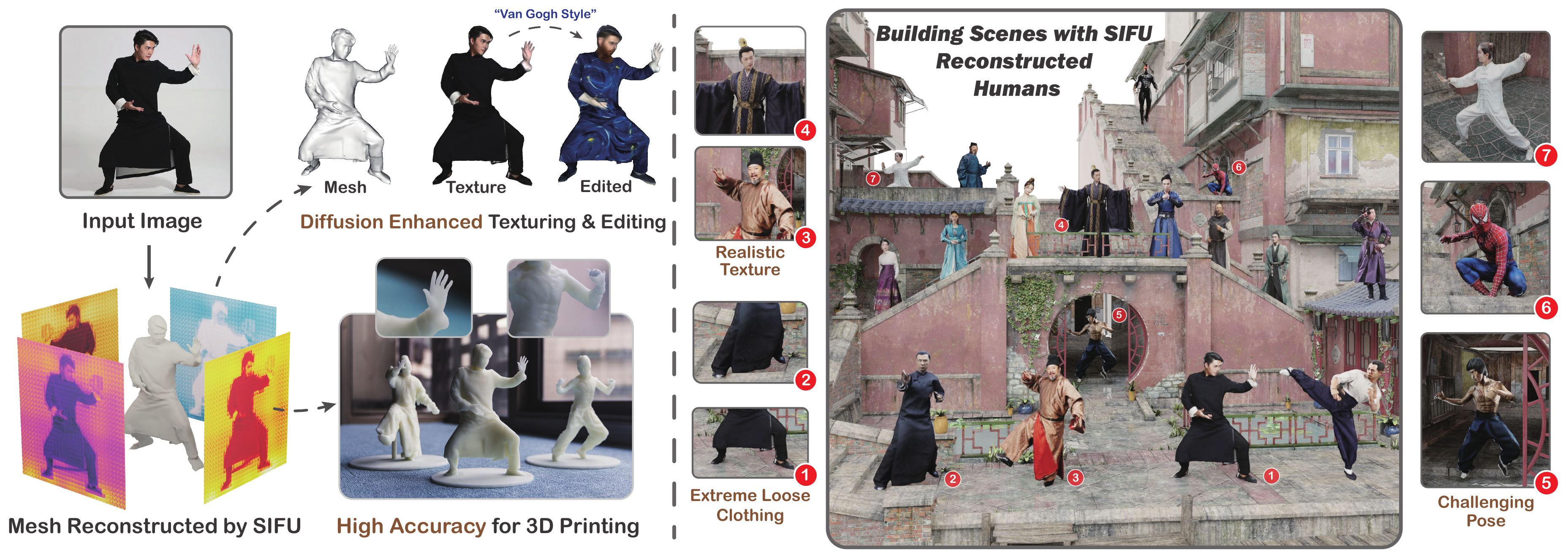}
    \end{minipage}
      \vspace{-0.5 em}
    \captionsetup{type=figure}
    \captionof{figure}{\looseness=-1{\teaserCaption}}
    \label{fig:teaser}
    \vspace{2.0 em}
}]


\renewcommand{\thefootnote}{\fnsymbol{footnote}}
\footnotetext[2]{\footnotesize Zongxin Yang is the corresponding author.}
\renewcommand*{\thefootnote}{\arabic{footnote}}

\begin{abstract}
Creating high-quality 3D models of clothed humans from single images for real-world applications is crucial. Despite recent advancements, accurately reconstructing humans in complex poses or with loose clothing from in-the-wild images, along with predicting textures for unseen areas, remains a significant challenge. A key limitation of previous methods is their insufficient prior guidance in transitioning from 2D to 3D and in texture prediction.
In response, we introduce \textbf{\mn} (\textbf{S}ide-view Conditioned \textbf{I}mplicit \textbf{F}unction for Real-world \textbf{U}sable Clothed Human Reconstruction), a novel approach combining a Side-view Decoupling Transformer with a 3D Consistent Texture Refinement pipeline.
\mn employs a cross-attention mechanism within the transformer, using SMPL-X normals as queries to effectively decouple side-view features in the process of mapping 2D features to 3D. This method not only improves the precision of the 3D models but also their robustness, especially when SMPL-X estimates are not perfect. Our texture refinement process leverages text-to-image diffusion-based prior to generate realistic and consistent textures for invisible views. Through extensive experiments, \mn surpasses SOTA methods in both geometry and texture reconstruction, showcasing enhanced robustness in complex scenarios and achieving an unprecedented Chamfer and P2S measurement. Our approach extends to practical applications such as 3D printing and scene building, demonstrating its broad utility in real-world scenarios. 
\end{abstract}

\vspace{-1.0 em}

\section{Introduction}
\label{sec:intro}

\begin{figure}[t]
\centering
\scriptsize
\includegraphics[width=\linewidth]{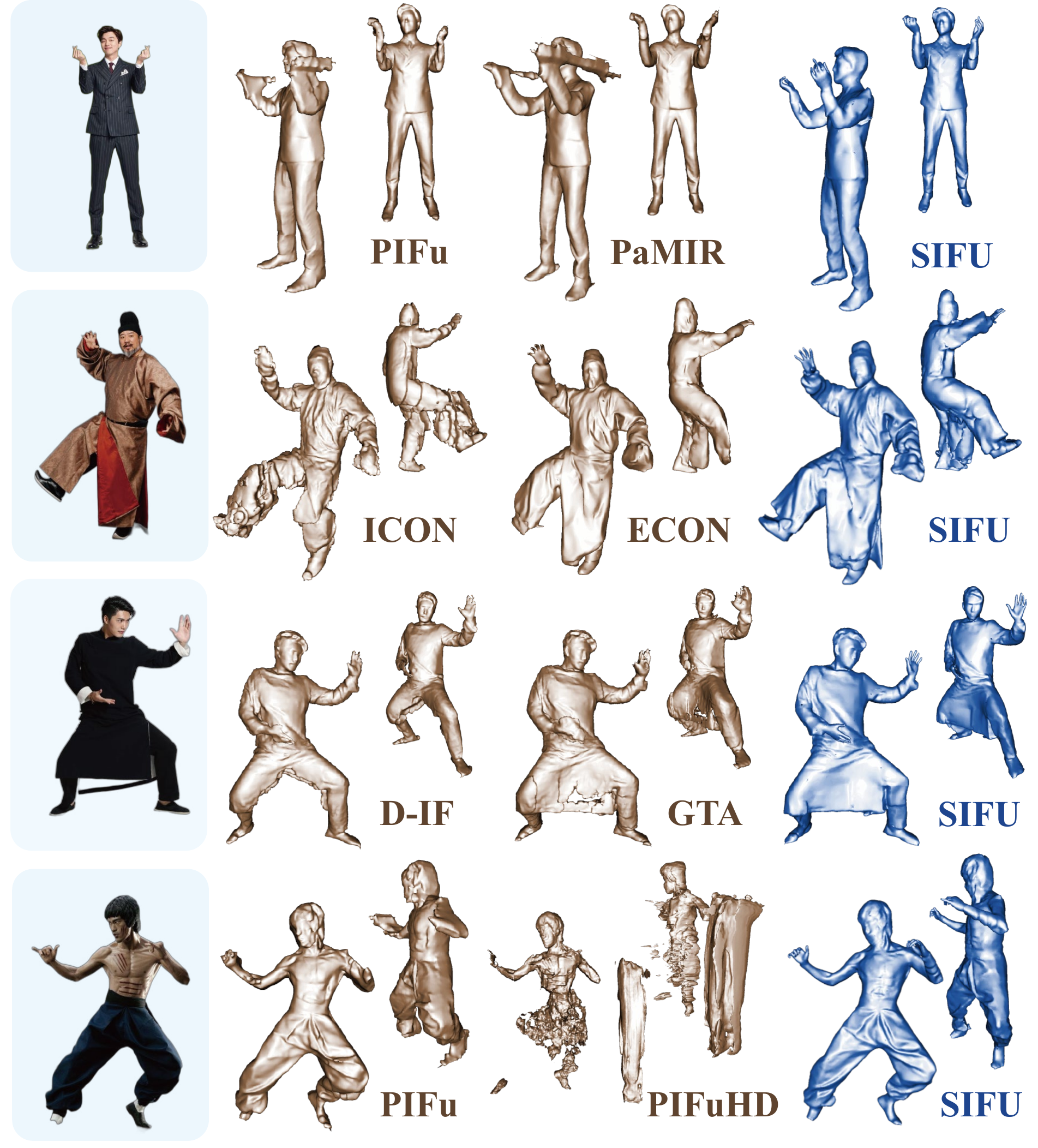}  
 \vspace{-2.0 em}
\scriptsize
\caption{\textbf{Comparison of SIFU with State-of-the-Art (SOTA) Methods in 3D Human Inference from In-the-Wild Images.} Existing SOTA methods often struggle with complex poses and loose clothing, leading to a range of artifacts. These issues include the absence of human shapes (\textcolor{Sepia}{PIFu}, \textcolor{Sepia}{PaMIR}, \textcolor{Sepia}{PIFuHD}), missing body parts (\textcolor{Sepia}{ECON}), disrupted clothing (\textcolor{Sepia}{ICON}, \textcolor{Sepia}{D-IF}), and a lack of fine details (\textcolor{Sepia}{GTA}). In contrast, \textcolor{RoyalBlue}{SIFU} effectively addresses these challenges, delivering high-quality, detailed results.}
\vspace{-1.0 em}
\label{fig:vsSOTA}
\end{figure}

High-quality 3D models of clothed humans are crucial in diverse sectors, including augmented and virtual reality (AR/VR), 3D printing, scene assembly, and filmmaking. The traditional process of creating these models not only requires a considerable amount of time but also specialized equipment capable of capturing multi-view photographs, in addition to the reliance on skilled artists. Contrasting this, in everyday situations, we most often have access to monocular images of individuals, easily obtained through phone cameras or found on various web pages. Thus, a method that accurately reconstructs 3D human models from a single image could significantly cut costs and simplify the process of independent creation. While existing deep learning models~\cite{saito2019pifu,Huang:ARCH:2020,saito2020pifuhd,zheng2021pamir,xiu2022icon,xiu2022econ,yang2023dif,zhang2023globalcorrelated,alldieck2022photorealistic,corona2023s3f} show promise in this area, they struggle with complex poses and loose clothing, as illustrated in~\cref{fig:vsSOTA}. Furthermore, these models fail to correctly texture hidden areas, resulting in less realistic outcomes. Therefore, there's a significant need for models that can generalize across various scenarios and efficiently produce realistic, real-world applicable 3D clothed humans. 

\begin{figure}[t]
\centering
\scriptsize
\includegraphics[width=\linewidth]{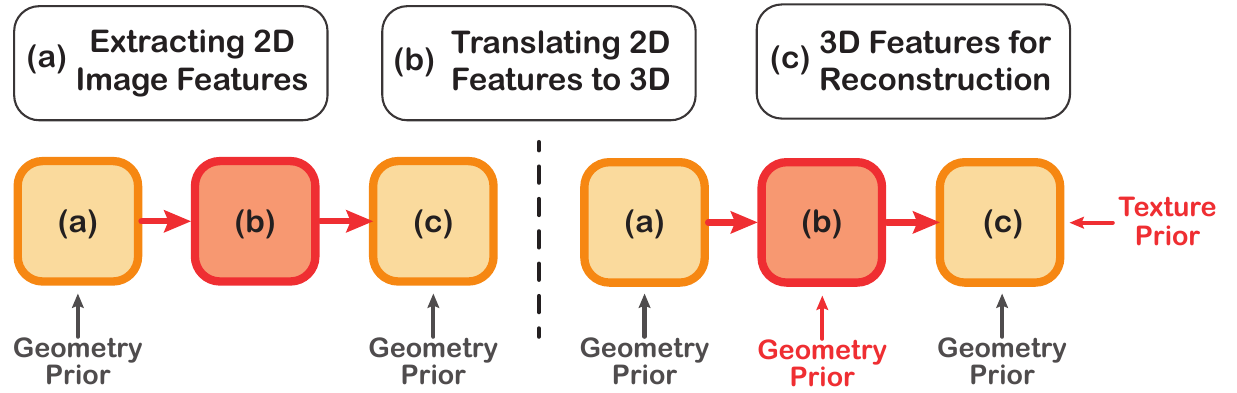}  
\vspace{-2.0 em}
\scriptsize
\caption{\textbf{Contrast between previous methods (Left) and ours (Right):} Our approach improves the reconstruction process by incorporating additional guidance on geometry and texture priors.
 }
\vspace{-1.0 em}
\label{fig:high_level}
\end{figure} 

Through analyzing existing methods, we pinpointed two key challenges in this field:
\textbf{(i) Insufficient Prior Guidance in Translating 2D Features to 3D:} The reconstruction of 3D objects from 2D images typically involves three main steps: (a) \textit{extracting 2D image features}, (b) \textit{translating 2D features to 3D}, and (c) \textit{3D features for reconstruction}. As shown by~\cref{fig:high_level}, current approaches often add geometric prior (like SMPL-X~\cite{SMPL-X:2019}) to the first and last steps, focusing on techniques such as normal map prediction~\cite{xiu2022icon,xiu2022econ,yang2023dif}, SMPL-guided SDF~\cite{xiu2022icon,zhang2023globalcorrelated,yang2023dif}, or volume features~\cite{zheng2021pamir,cao2023sesdf}. While the use of priors for improving the transition from 2D image features to 3D is crucial, it remains underexplored. Currently, this transition is typically achieved by projecting features onto 3D points~\cite{saito2019pifu,saito2020pifuhd,xiu2022icon,zheng2021pamir,corona2023s3f,alldieck2022photorealistic,yang2023dif,cao2023sesdf} or by employing fixed learnable embeddings to generate 3D features~\cite{zhang2023globalcorrelated}. These methods, however, do not fully harness the potential of priors in enhancing accuracy of 3D reconstruction. \textbf{(ii) Lack of Texture Prior:} While current methods~\cite{saito2019pifu,saito2020pifuhd,alldieck2022photorealistic,corona2023s3f,zhang2023globalcorrelated} attempt to predict vertex colors, they struggle to accurately predict textures for unseen views, particularly with limited training data. This limitation highlights a need for additional texture priors in 3D human reconstruction.

In response to the challenges we've identified, we propose two refined strategies to enhance 3D human reconstruction. \textbf{Firstly}, we believe that enhancing the process of translating 2D features to 3D with additional guidance could significantly improve both the accuracy and efficiency of 3D reconstructions. To more effectively integrate prior guidance, such as SMPL-X~\cite{SMPL-X:2019}, with image features, we utilize the cross-attention mechanism of the transformer~\cite{Transformer:2017}. This approach aims to optimize the fusion of geometry and image data, potentially leading to more precise and realistic 3D human models.
\textbf{Secondly}, considering the impressive generative capabilities of pretrained diffusion models, as shown in recent studies~\cite{sohl2015deepdiffu,ho2020denoisingdiffu,dhariwal2021diffusion,croitoru2023diffusion,nichol2021improveddiffu} and their proficiency in learning rich 3D priors~\cite{liu2023zero1to3,liu2023one2345,qian2023magic123,lin2023magic3d,tang2023make-it-3d,liu2023syncdreamer,poole2022dreamfusion}, we suggest their incorporation as priors to enhance texture prediction, particularly for invisible regions. Besides, maintaining 3D consistency from different angles and matching the style of the input image is also crucial for creating realistic textures.

In this paper, we present \textbf{\mn} (\textbf{S}ide-view Conditioned \textbf{I}mplicit \textbf{F}unction for Real-world \textbf{U}sable Clothed Human Reconstruction), a novel approach employing a \textbf{Side-view Conditioned Implicit Function (\S\ref{sec:side-view implicit})} with a \textbf{3D Consistent Texture Refinement (\S\ref{sec:texture refinement})} pipeline for precise geometry and realistic texture reconstruction. Our approach employs normals from SMPL-X as queries in a cross-attention mechanism with image features. This method effectively decouples side-view features in the process of mapping 2D features to 3D, thereby enhancing the accuracy and robustness of reconstruction. Moreover, our texture refinement employs text-to-image diffusion models~\cite{rombach2021highresolution} and ensures uniform diffusion features across different perspectives, resulting in detailed, consistently styled textures.

Through extensive experiments, \textbf{SIFU} outperforms existing SOTA methods in geometry and texture quality, achieving an unprecedented Chamfer and P2S measurement of \textbf{0.6 cm} on THuman2.0~\cite{THuman2.0:2021} (\tab{table:geo-metrics}). Additionally, SIFU shows improved robustness in geometry reconstruction (\tab{table:ablation-smpl}), even with inaccurate SMPL-X estimations. SIFU handles complex poses and loose clothing well, producing realistic textures with consistent colors and patterns (\cref{fig: qualitative figure}). Its adaptability extends to practical applications like 3D printing and scene creation (\cref{fig:teaser}), showcasing its broad practical utility. Key contributions include:

\begin{itemize}
\item  A novel \textbf{Side-view Conditioned Implicit Function} that skillfully maps 2D image features to 3D with SMPL-X guidance. This is the first instance showcasing the efficacy of using human prior information to decouple side-view 3D features from the input image, significantly advancing the field of clothed human reconstruction.

\item A \textbf{3D Consistent Texture Refinement} pipeline designed to generate realistic, 3D consistent textures on clothed human meshes. This approach has notably improved the quality and uniformity of textures, offering a substantial advancement in the field.
\item Our proposed model achieves state-of-the-art performance in both geometry and texture reconstruction, facilitating \textbf{real-world applications} such as 3D printing and scene building, which were challenging to achieve with previous methods.
\end{itemize}

\section{Related Work}
\label{sec:related}

\noindent \textbf{Implicit-function-based Reconstruction.} Implicit representations, such as occupancy and signed distance fields, are flexible with topology and can effectively depict 3D clothed humans across a variety of scenarios, including loose garments and complex poses. A series of studies have focused on regressing the implicit surface from a single input image directly in a streamlined process~\cite{saito2019pifu,saito2020pifuhd,alldieck2022photorealistic,han2023high,albahar2023humansgd}. Others incorporate a 3D human body prior to enhance the process of 2D feature extraction and 3D feature for reconstruction~\cite{he2020geoPifu,corona2023s3f,xiu2022icon,zheng2021pamir,huangARCHAnimatableReconstruction2020,heARCHAnimationReadyClothed2021,cao2022jiff,liao2023car,yang2023dif,cao2023sesdf,zhang2023globalcorrelated,huang2024tech,xiu2022econ}. Among these, GTA~\cite{zhang2023globalcorrelated} utilizes transformers with fixed learnable embeddings to translating image features to 3D tri-plane features. As for texture reconstruction, methods like \pifu~\cite{saito2019pifu}, ARCH~\cite{huangARCHAnimatableReconstruction2020,heARCHAnimationReadyClothed2021}, \pamir~\cite{zheng2021pamir}, and GTA~\cite{zhang2023globalcorrelated} deduce full textures from a single image. Techniques such as \phorhum~\cite{alldieck2022photorealistic} and S3F~\cite{corona2023s3f} go further by segregating albedo and global illumination. Nevertheless, these methods lack information from other views or prior knowledge (such as diffusion models), resulting in unsatisfactory textures. HumanSGD~\cite{albahar2023humansgd} employs diffusion models for mesh inpainting but faces performance declines with mesh reconstruction inaccuracies. TeCH~\cite{huang2024tech} uses diffusion-based models for visualizing unseen areas, yielding realistic results. Its limitations, however, include time-intensive per-subject optimization and dependence on accurate SMPL-X.

\noindent \textbf{Explicit-shape-based Reconstruction.} Recovering the human mesh from a single RGB image is a complex challenge that has received extensive attention. Many approaches~\cite{pymaf2021, PIXIE:2021, li2022cliff, li2021hybrik, Kanazawa2018_hmr, pare,spec,pymafx2023,li2023niki,VIBE:CVPR:2020,Kolotouros2019_spin,feng2023learning} adopt parametric body models~\cite{SMPL:2015, xu2020ghum, pavlakos2019expressive,Joo_2018_CVPR} to estimate the shape and pose of a 3D human body with minimal clothing~\cite{shen2023global,li2023jotr}. To incorporate clothing into the 3D models, methods often apply 3D clothing offsets~\cite{alldieckTex2ShapeDetailedFull2019, alldieck2018dv,alldieck2018videoavatar,alldieck2019peopleInClothing,zhu2019hierarchMeshDeform,lazova2019textures360,xiang2020monoClothCap} or use adjustable garment templates~\cite{bhatnagar2019multiGarmentNet,jiang2020bcnet,feng2023learning} over the base body shape. Additionally, non-parametric forms like depth maps~\cite{smith2019facsimile,gabeur2019moulding}, normal maps~\cite{xiu2022econ}, and point clouds~\cite{zakharkin2021point} are explored for creating representations of clothed humans.

Despite these advancements, explicit-shape approaches can be limited by topological constraints, which become apparent when handling diverse and complex clothing styles found in real-world settings, such as dresses, and skirts.

\noindent \textbf{NeRF-based Reconstruction.} The rise of Neural Radiance Fields (NeRF) has seen methods~\cite{Peng_2021_CVPR,Peng_2021_ICCV,Pan_2023_ICCV,jiang2022instantavatar,mu2023actorsnerf,weng_humannerf_2022_cvpr,instant_nvr,yu2023monohuman,jiang2022neuman,guo2023vid2avatar} using videos or multi-view images to optimize NeRF for human form capture. Recent advancements like SHERF~\cite{hu2023sherf} and ELICIT~\cite{huang2022elicit} aim to generate human NeRFs from single images, with SHERF filling gaps using 2D image data and ELICIT employing a pre-trained CLIP model~\cite{clip} for contextual understanding. While NeRF-based approaches are effective in creating quality images from various perspectives, they typically struggle with detailed 3D mesh generation from single images and often require extensive time for optimization.

Contrasting with these methods, \mn stands out in reconstructing clothed human meshes across various scenarios, producing consistently realistic 3D textures suitable for real-world use. It leverages human body priors to decouple side-view features from input images during the 2D to 3D mapping process, thereby improving the accuracy of its implicit function. For texture refinement, \mn adopts a coarse-to-fine approach, utilizing a pre-trained diffusion model, trained on a vast dataset, to predict textures in unseen areas. It also reconstructs texture from the input image for visible regions, ensuring uniform texture consistency.

\newcommand{\pipelineCaption}{
Given a single image, \mn constructs a 3D clothed human mesh with coarse textures using a \textbf{Side-view Conditioned Implicit Function} (\S\ref{sec:side-view implicit}). This is followed by a step of \textbf{3D Consistent Texture Refinement} (\S\ref{sec:texture refinement}) to generate detailed textures. Specifically, \mn employs a side-view decoupling transformer to decouple features from the input image and the side-view normals of the SMPL-X model. Then, these features are combined at a query point through a hybrid prior fusion strategy, aiding in the reconstruction of both the mesh and its texture. Finally, the mesh with its basic textures undergoes a diffusion-based 3D consistent texture refinement, ensuring feature consistency in the latent space and resulting in high-quality textures.}


\begin{figure*}[tbp]
    \centering
    \scriptsize
    \includegraphics[width=\linewidth]{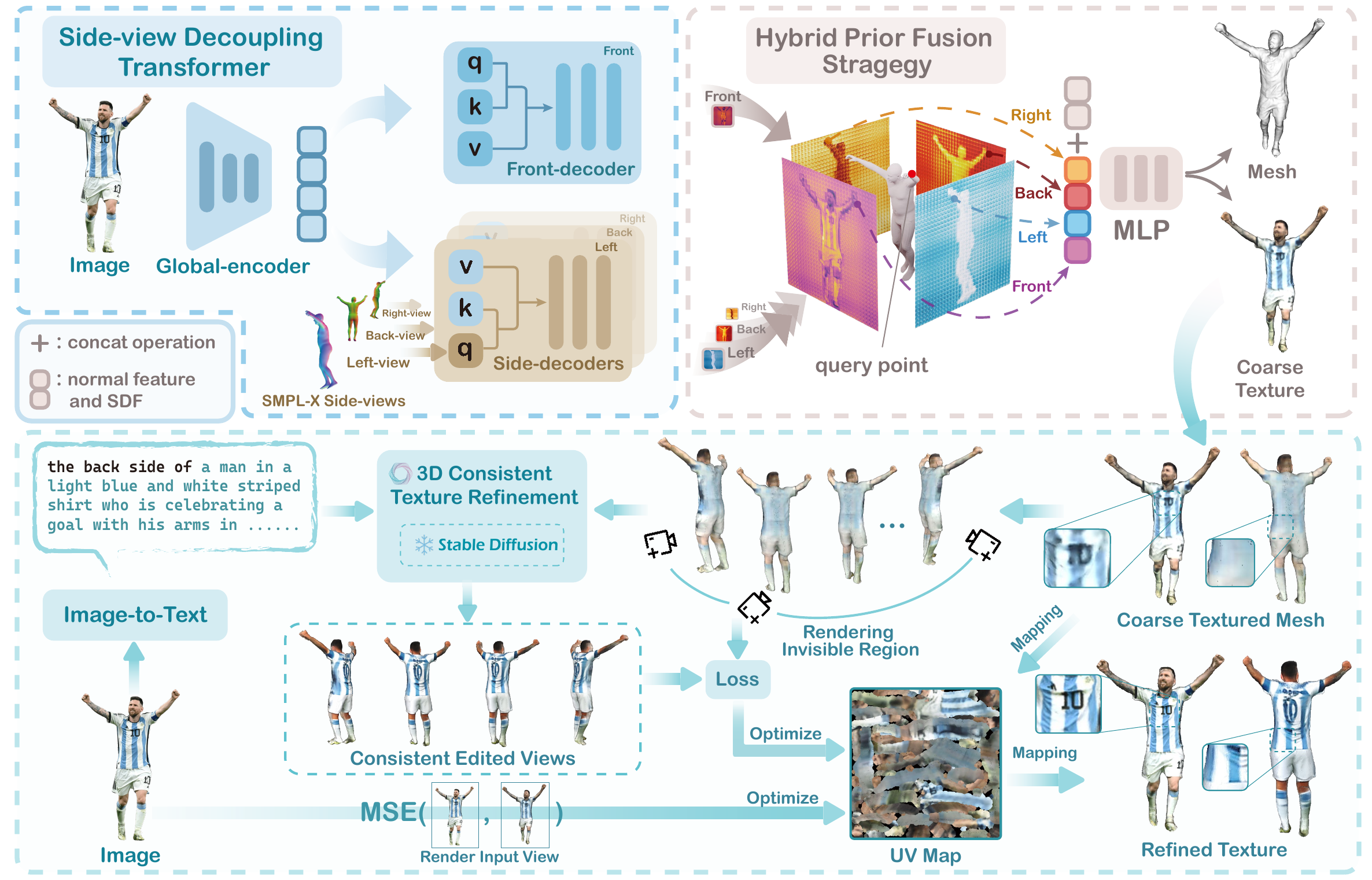}
    \vspace{-1.5 em}
    \captionof{figure}{\pipelineCaption}
     \vspace{-1.5 em}
    \label{fig:pipeline}
\end{figure*}

\section{Method}\label{sec:method}
Given a single image, \mn first reconstructs the 3D mesh and coarse textures using the Side-view Conditioned Implicit Function (\sect{sec:side-view implicit}). Subsequently, it employs a 3D Consistent Texture Refinement process (\sect{sec:texture refinement}) to enhance textures, ensuring high quality and 3D consistency. Key preliminary concepts necessary for understanding our approach are briefly presented in~\sect{sec:preliminary}.

\subsection{Preliminary}\label{sec:preliminary}

\textbf{Implicit Function} is a powerful tool for modeling complex geometries and colors with neural networks. We employ implicit function to predict an occupancy field to represent 3D clothed humans. Specifically, our implicit function $\mathcal{IF}$ maps an input point $\boldsymbol{x}$ to a scalar value representing the spatial field including occupancy and color fields. Our reconstructed human surface can be represented as $\mathcal{S}_\mathcal{IF}$:
\begin{equation}
    \mathcal{S}_\mathcal{IF} = \{\boldsymbol{x} \in \mathbb{R}^3 \mid \mathcal{IF}(\boldsymbol{x}) = (\boldsymbol{o}, \boldsymbol{c})\}
\end{equation}
where occupancy $\boldsymbol{o}$ = 0.5 and color $\boldsymbol{c}\in \mathbb{R}^3$.

\noindent\textbf{SMPL and SMPL-X.} The Skinned Multi-Person Linear (SMPL) model~\cite{SMPL:2015} is a parametric model for human body representation. It uses shape parameters $\boldsymbol{\beta} \in \mathbb{R}^{10}$ and pose parameters $\boldsymbol{\theta} \in \mathbb{R}^{3\times 24}$ to define the human body mesh $\mathcal{M}$:
\begin{equation}
\mathcal{M}(\boldsymbol{\beta},\boldsymbol{\theta}) : \boldsymbol{\beta} \times \boldsymbol{\theta} \mapsto \mathbb{R}^{3 \times 6890}
\end{equation}
Here, $\boldsymbol{\beta}$ controls body size, while $\boldsymbol{\theta}$ affects joint positions and orientations. The SMPL-X model~\cite{SMPL-X:2019} builds upon SMPL, adding features for hands and face, enhancing facial expressions, finger movements, and detailed body poses.

\noindent\textbf{Diffusion Models.} Diffusion processes, notably represented by Diffusion Probabilistic Models (DPM) \cite{sohl2015deepdiffu,ho2020denoisingdiffu,dhariwal2021diffusion,croitoru2023diffusion,nichol2021improveddiffu}, are pivotal in image generation and have shown capabilities in human/avatar generation~\cite{huang2023avatarfusion,xu2023seeavatar}. These models aim to approximate a data distribution \(q\) through a progressive denoising process. Starting with a Gaussian i.i.d noisy image \(\boldsymbol{x}_T\sim \mathcal{N}(0,I)\), the model denoises it until a clean image \(\boldsymbol{x}_0\) from the target distribution \(q\) is obtained. DPMs can also learn a conditional distribution with additional guiding signals like text conditioning.


\subsection{Side-view Conditioned Implicit Function}\label{sec:side-view implicit}

The Side-view Conditioned Implicit Function in our model comprises two key components: the \textbf{Side-view Decoupling Transformer} and the \textbf{Hybrid Prior Fusion Strategy}. The transformer initially uses rendered SMPL-X images from various side views as queries to perform cross-attention with the encoded input image. This process effectively decouples features conditioned on the side views. The Hybrid Prior Fusion Strategy then integrates these features at each query point, which are later input into a Multi-Layer Perceptron (MLP) for predicting occupancy and color. We detail both components in the sections below.

\noindent \textbf{Side-view Decoupling Transformer.} Our method draws inspiration from the shared characteristics, such as material and color, between side views (like the back or left side) and the visible front view. Despite their different perspectives, these similarities in features are crucial. Therefore, we aim to effectively separate side-view features from the front view, utilizing the SMPL-X model~\cite{SMPL-X:2019} as a guide.

The process begins with a ViT-based global encoder~\cite{ViT:2020}, which encodes the input image $I$ into a latent feature $h$, capturing the image's globally correlated features. To decode these features, we employ two decoders: a front-view decoder, aligned with $h$, and a side-view decoder. The front-view decoder utilizes multi-head self-attention within a vision transformer to process the front view feature, represented as $F_{front}\in \mathbb{R}^{H\times W\times C}$.


To decouple side-view features, we render the side-view normal images $N_i$ of SMPL-X as guidance, with $i\in \{left, \ back,\ right\}$ during the experiments. The side-view normals $N_i$ are transformed to embeddings $z_i$, which then engage in a cross-attention operation as queries, with the latent feature $h$ acting as both keys and values:
\begin{equation}
    \small
    \mathrm{\mathbf{CrossAttn}}(\boldsymbol{z_i},\boldsymbol{h})=\mathrm{\mathbf{SM}}(\frac{(W^Q\boldsymbol{z_i})(W^K\boldsymbol{h})^T)}{\sqrt{d}})(W^V\boldsymbol{h})
\end{equation}
\noindent where $\mathbf{SM}$ represents $\mathbf{SoftMax}$ operation, while $W^Q$, $W^K$, and $W^V$ are learnable parameters and $d$ is the scaling coefficient. Following the original transformer architecture~\cite{Transformer:2017}, our model integrates residual connections~\cite{he2016deepresiduallearning} and layer normalization~\cite{ba2016layer} after each sub-layer. The entire side-decoder contains multiple identical layers, and we deploy three such decoders to yield feature maps $F_{i}\in \mathbb{R}^{H\times W\times C}$ where $i \in \{left,\ back,\ right\}$.

\noindent \textbf{Hybrid Prior Fusion Strategy.} In our pipeline, we incorporate the Hybrid Prior Fusion Strategy from~\cite{zhang2023globalcorrelated} to effectively merge features at a query point, utilizing both spatial localization and human body prior knowledge. We split the feature maps $F_j$ (for $j\in \{front, left, back, right\}$) into two groups. For the spatial query group, we project query points onto the feature map to obtain pixel-aligned features $F^{S}_j$. We then combine these features from all planes using a mix of averaging and concatenation:
\begin{equation}
    F^{S}(\boldsymbol{x}) = F_{f}^{S}(\boldsymbol{x}) \oplus \mathbf{avg}(F_{l}^{S}(\boldsymbol{x}), F_{b}^{S}(\boldsymbol{x}),F_{r}^{S}(\boldsymbol{x}))
\end{equation}
\noindent where $f,\ l,\ b,\ r$ denote the front, left, back, and right respectively. For the other group, similar to the spatial query, we project the SMPL-X~\cite{SMPL-X:2019} mesh vertices onto the four feature maps, obtaining the feature $F^S(\boldsymbol{v})$, $\boldsymbol{v}\in \mathcal{M}$, where $\mathcal{M}$ is the SMPL-X mesh. For each query point $\boldsymbol{x}$, we find its nearest triangular face $t_{\boldsymbol{x}}=[\boldsymbol{v}_0, \boldsymbol{v}_1, \boldsymbol{v}_2] \in \mathbb{R}^{3\times3}$ and employ barycentric interpolation to integrate features for $\boldsymbol{x}$, denoted as $F^P(\boldsymbol{x})$:
\begin{equation}
    F^{P}(\boldsymbol{x}) = uF^{S}(\boldsymbol{v}_0) + vF^{S}(\boldsymbol{v}_1) + wF^{S}(\boldsymbol{v}_2)
\end{equation}
\noindent where $[u, v, w]$ represents the barycentric coordinates of the query point $\boldsymbol{x}$ projected onto triangle $t_{\boldsymbol{x}}$. We concatenate these two query features as the final point feature. Moreover, we incorporate the signed distance between the query point and SMPL-X mesh $\mathcal{SDF}(\boldsymbol{x})$ and pixel-aligned normal feature $F^{N}(\boldsymbol{x})$ as input to a Multilayer Perceptron (MLP) for prediction of occupancy and color:
\begin{equation}
    \small
    (\boldsymbol{o},\boldsymbol{c}) = \mathbf{MLP}(F^{S}(\boldsymbol{x}),F^{P}(\boldsymbol{x}),\mathcal{SDF}(\boldsymbol{x}),F^{N}(\boldsymbol{x}))
\end{equation}
\noindent \textbf{Training Objectives.} We consider two sets of points as training data, denoted as $G_o$ and $G_c$. $G_c$ is sampled uniformly with a slight perturbation along the normals of the ground-truth mesh surface, whereas $G_o$ is sampled according to the same strategy as in~\cite{saito2019pifu}. For the points in $G_o$, we employ the following loss function:
\begin{equation}
    \small
    \mathcal{L}_o=\frac{1}{|G_o|}\sum_{\boldsymbol{x}\in G_o}BCE(\hat{\mathit{o}}_{\boldsymbol{x}}-\mathit{o}_{\boldsymbol{x}})
\end{equation}

\noindent where $\hat{\mathit{o}}_{\boldsymbol{x}}$ denotes the model's predicted occupancy, while $\mathit{o}_{\boldsymbol{x}}$ is the ground-truth occupancy. For the sampled points in $G_c$, we apply the following loss function:
\begin{equation}
    \small
    \mathcal{L}_c=\frac{1}{|G_c|}\sum_{\boldsymbol{x}\in G_c}|\boldsymbol{\hat c_x}-\boldsymbol{c_x}|
\end{equation}

\noindent where $\boldsymbol{\hat c_x}$ denotes the predicted color, and $\boldsymbol{c_x}$ represents the corresponding ground-truth color. The total loss is the sum of these two separate losses, which is designed to fulfill a comprehensive training objective.

\noindent\textbf{Mesh Extraction.} We begin by densely sampling points in space and using our side-view conditioned implicit function to predict their occupancy values. The Marching Cubes algorithm~\cite{lorensen1987marching} is then applied to extract the mesh, and following~\cite{xiu2022econ}, we substitute the hands with SMPL-X models for enhanced visuals. Finally, these mesh points are processed through the implicit function again for color prediction.

\begin{table*}[t]
\centering
\footnotesize
\resizebox{\linewidth}{!}{
\begin{tabular}{rcccccccccc}
\shline
\rowcolor{gray!30} 
\multicolumn{2}{c|}{} &
  \multicolumn{3}{c|}{CAPE-NFP} &
  \multicolumn{3}{c|}{CAPE-FP} &
  \multicolumn{3}{c}{\thtwo} \\
  \rowcolor{gray!30} 
  Method & \multicolumn{1}{r|}{Publication} &
  Chamfer $\downarrow$ &
  P2S $\downarrow$ &
  \multicolumn{1}{c|}{Normal $\downarrow$} &
  Chamfer $\downarrow$ &
  P2S $\downarrow$ &
  \multicolumn{1}{c|}{Normal $\downarrow$} &
   Chamfer $\downarrow$ &
  P2S $\downarrow$ &
  \multicolumn{1}{c}{Normal $\downarrow$} \\ \shline
\multicolumn{11}{c}{\textit{w/o SMPL-X body prior}} \\ \hline
 \pifu*~\cite{saito2019pifu} & 
\multicolumn{1}{r|}{ICCV 2019} &
 2.5609  &
  1.9971 &
  \multicolumn{1}{c|}{0.1023} &
  1.8139 &
  1.5108 &
  \multicolumn{1}{c|}{0.0798} & 
  1.5991 &
  1.4333 &
  \multicolumn{1}{c}{0.0843}   \\
\pifuhd~\cite{saito2020pifuhd} &
\multicolumn{1}{r|}{CVPR 2020} &
  3.7670 &
  3.5910 &
  \multicolumn{1}{c|}{0.1230} &
 2.3020  &
  2.3350 &
  \multicolumn{1}{c|}{0.0900} &
  - &
  - &
  \multicolumn{1}{c}{-}  \\ \hline
\multicolumn{11}{c}{\textit{w/ \smplx body prior}} \\ \hline
\pamir*~\cite{zheng2021pamir} &
\multicolumn{1}{r|}{TPAMI 2021} &
  1.6313 &
  1.2666 &
  \multicolumn{1}{c|}{0.0730} &
  1.481 &
  1.1631&
  \multicolumn{1}{c|}{0.0727} & 
  1.2152 &
  1.0582 &
  \multicolumn{1}{c}{0.0730} \\
\icon~\cite{xiu2022icon} & 
\multicolumn{1}{r|}{CVPR 2022} &
  0.8846 &
  0.8569 &
  \multicolumn{1}{c|}{0.0434} & 
  0.7247 &
  0.6979 & 
  \multicolumn{1}{c|}{0.0371} &
  0.9491 &
  0.9846 &
  \multicolumn{1}{c}{0.0621}\\
\econ~\cite{xiu2022econ} &\multicolumn{1}{r|}{CVPR 2023} &
  0.9462 &
  0.9334 &
  \multicolumn{1}{c|}{\btwo 0.0382} & 
  0.9039 &
  0.8938 & 
  \multicolumn{1}{c|}{0.0373} &
  1.2585 &
  1.4184 &
  \multicolumn{1}{c}{0.0612} \\ 
  \dif~\cite{yang2023dif} & 
\multicolumn{1}{r|}{ICCV 2023} &
  \btwo 0.8237 &
  0.8353 &
  \multicolumn{1}{c|}{0.0575} & 
  0.7625 &
  0.769 & 
  \multicolumn{1}{c|}{0.0503} &
 
  1.1696 &
  1.2900 &
  \multicolumn{1}{c}{0.0936}\\
    \gta~\cite{zhang2023globalcorrelated} & 
\multicolumn{1}{r|}{NeurIPS 2023} &
  0.8508 &
  \btwo 0.7920 &
  \multicolumn{1}{c|}{0.0424} & 
 \btwo 0.6525 &
  \btwo 0.6084 & 
  \multicolumn{1}{c|}{\btwo 0.0349} &
 
  \btwo0.7329 &
  \btwo 0.7297 &
  \multicolumn{1}{c}{\btwo 0.0492}\\
  
  \hline
   
\textbf{Ours} &\multicolumn{1}{c|}{-} &

\bone \textbf{0.7725}
   & \bone \textbf{0.7354}
   &
  \multicolumn{1}{c|}{\bone \textbf{0.0378}} &
  \bone \textbf{0.6297} &
  \bone \textbf{0.5980} &

  \multicolumn{1}{c|}{\bone \textbf{0.0327}} &
  \bone \textbf{0.5961} &
  \bone \textbf{0.6058} &
  \multicolumn{1}{c}{\bone \textbf{0.0407}} \\ \shline
\end{tabular}}
 \vspace{-1.0 em}
\caption{\textbf{Quantitative evaluation against SOTA (\S\ref{sec:evaluation}). }
All models use a resolution of 256 for marching cubes and ground-truth SMPL-X models are used during testing. *Methods are re-implemented in~\cite{xiu2022icon} for a fair comparison. Top two results are colored as \colorbox{bestcolor}{first}~\colorbox{secondbestcolor}{second}.} 
\vspace{-1.0 em}
\label{table:geo-metrics}
\end{table*}

\subsection{3D Consistent Texture Refinement}\label{sec:texture refinement}

Upon extracting the mesh using our implicit function, we noted that color quality was coarse and areas not visible in the input were blurry, leading to a less realistic look (see~\cref{fig:pipeline}). To address this, we developed a \textbf{3D Consistent Texture Refinement} pipeline, leveraging text-to-image diffusion priors to substantially enhance texture quality.

\noindent \textbf{Pipeline.} For a given input image and its reconstructed mesh ${M}$, we first utilize vision-to-text models (\eg,~\cite{OpenAI_GPT4_2023,yang2024doraemongpt,ma2023vista}) to convert the image into a textual description ${P}$, and then back-project the mesh color onto a UV texture map ${{T}}$, following the approach in~\cite{tang2023dreamgaussian}. To visualize unseen mesh areas, differentiable rendering $\mathcal{I}$ is employed on mesh $M$, generating images of these invisible views:
\begin{equation}
    \boldsymbol{I}=\mathcal{I}({T},{M},\boldsymbol{k})
\end{equation}
where $\boldsymbol{k}=\{k^1,...,k^n\}$ represent camera views and $\boldsymbol{I}=\{I^1,...,I^n\}$ are the corresponding rendered images.

Subsequently, a pretrained and fixed text-to-image diffusion model $\boldsymbol{\epsilon_\theta}$ refines the blurry images $\boldsymbol{I}$ into enhanced images $\boldsymbol{J}$, using $P$ as a condition. To ensure consistency among refined images, a \textbf{consistent editing} technique $\mathcal{H}$ is applied to $\boldsymbol{\epsilon_\theta}$, preserving the original semantic layout of $\boldsymbol{I}$:
\begin{equation}
    \boldsymbol{J}=\mathcal{H}(\boldsymbol{\epsilon_\theta},P,\boldsymbol{I})=\mathcal{H}(\boldsymbol{\epsilon_\theta},P,\mathcal{I}({T},{M},\boldsymbol{k}))
\end{equation}
where $\boldsymbol{J}=\{J^1,...,J^n\}$ corresponds to the refined views of $\boldsymbol{I}$. After obtaining $\boldsymbol{J}$, a pixel-wise Mean Squared Error (MSE) loss is computed between each $J^i$ and $I^i$ to optimize the texture map $T$. Additional losses include a perceptual loss $\mathcal{L}_{vgg}$~\cite{DBLP:journals/corr/JohnsonAL16} and a Chamfer Distance loss $\mathcal{L}_{CD}$~\cite{huang2024tech}, aimed at ensuring style similarity between $\boldsymbol{I}$ and the input image. We also compute an MSE loss $\mathcal{L}^f_{MSE}$ from the input view against the input image. These combined losses jointly optimize $T$, enhancing overall texture quality:
\begin{equation}
    \min_{T}\  \lambda_1\mathcal{L}_{MSE}+\lambda_2\mathcal{L}_{vgg}+\lambda_3\mathcal{L}_{CD}+\lambda_4\mathcal{L}^f_{MSE}
\end{equation}
where $\lambda_1,\lambda_2,\lambda_3,\lambda_4$ are the weights attributed to each loss.

\noindent\textbf{Consistent Editing.} To achieve consistent image editing across different views, we adopt a method inspired by~\cite{geyer2023tokenflow}. This involves enforcing consistency among diffusion features from various rendered views. We perform DDIM inversion~\cite{song2020_ddim} on the input image $\boldsymbol{I}$, extracting diffusion tokens across all layers. A set of key views is selected for joint editing~\cite{wu2023tune}, ensuring a unified appearance in the resultant features. These features are then propagated to all views using a nearest-neighbor approach to maintain coherence across them. Please refer to the SupMat for more detailed procedural insights and specific mechanisms.


\section{Experiment} \label{sec:experiment}

\textbf{Datasets.} We trained our model on the THuman2.0 dataset~\cite{THuman2.0:2021}, comprising 526 human scans, with 490 used for training, 15 for validation, and 21 for testing. Ground-truth SMPL-X models were used during training, and PIXIE~\cite{PIXIE:2021} was employed for inference. Our main evaluations were conducted on the CAPE~\cite{CAPE:2020} and THuman2.0 datasets. To test our model's versatility with different poses, we divided the CAPE dataset into "CAPE-FP" and "CAPE-NFP" subsets. Further details on datasets and implementation are available in the SupMat.

\subsection{Evaluation}\label{sec:evaluation}

\noindent\textbf{Metrics.} Our model's reconstruction quality for geometry is quantitatively evaluated using Chamfer and P2S distances, comparing reconstructed meshes with ground-truth. We also measure L2 Normal error between normal images from both meshes, assessing surface detail consistency by rotating the camera at $\{0^{\circ}, 90^{\circ}, 180^{\circ}, 270^{\circ}\}$ relative to the input view. For texture quality, we report the PSNR on colored images rendered similarly to normal images.

\begin{table}[t]
\centering
\scriptsize
\resizebox{\linewidth}{!}{
\begin{tabular}{rcccc}
\shline
\rowcolor{gray!30} 
 \multicolumn{1}{c|}{Method} &\multicolumn{1}{c|}{Backbone} &
  Chamfer $\downarrow$ &
  P2S $\downarrow$ &
  \multicolumn{1}{c}{Normal $\downarrow$} \\ \shline
\multicolumn{1}{r|}{\pamir~\cite{zheng2021pamir}}&
\multicolumn{1}{c|}{CNN}&
  1.3224 &
  1.1349 &
  \multicolumn{1}{c}{0.0767} \\
\multicolumn{1}{r|}{\icon~\cite{xiu2022icon}} & 
\multicolumn{1}{c|}{CNN}&
  1.2935 &
  1.3949 &
  \multicolumn{1}{c}{0.0781}\\
  \multicolumn{1}{r|}{\dif~\cite{yang2023dif}} & 
  \multicolumn{1}{c|}{CNN}&
  1.5262 &
  1.7296 &
  \multicolumn{1}{c}{0.1191}\\
\multicolumn{1}{r|}{\econ~\cite{xiu2022econ}} &
\multicolumn{1}{c|}{-}&
  2.1195 &
  1.8074 &
  \multicolumn{1}{c}{0.1029} \\ 
  
\multicolumn{1}{r|}{\gta~\cite{zhang2023globalcorrelated}} & 
\multicolumn{1}{c|}{Transformer}&
  \btwo 1.0473 &
  \btwo 1.0780 &
  \multicolumn{1}{c}{\btwo 0.0649}\\
  
  \hline
   
\multicolumn{1}{r|}{\textbf{Ours}} &
\multicolumn{1}{c|}{Transformer}&
  \bone \textbf{0.9937} &
  \bone \textbf{1.0645} &
  \multicolumn{1}{c}{\bone \textbf{0.0599}} \\ \shline
\end{tabular}}
 \vspace{-1.0 em}
\caption{\textbf{Assessing model robustness to SMPL-X (\S\ref{sec:evaluation}).}
To evaluate the models' robustness in reconstruction, we used the THuman2.0 dataset~\cite{THuman2.0:2021} and introduced random noise to the ground-truth SMPL-X models. This approach simulates inaccuracies in poses and shapes for robustness testing.}
\vspace{-2.0 em}
\label{table:ablation-smpl}
\end{table}


\begin{figure}[htbp]

\centering{
    \begin{subfigure}{\linewidth}
        \includegraphics[width=1.0\linewidth]{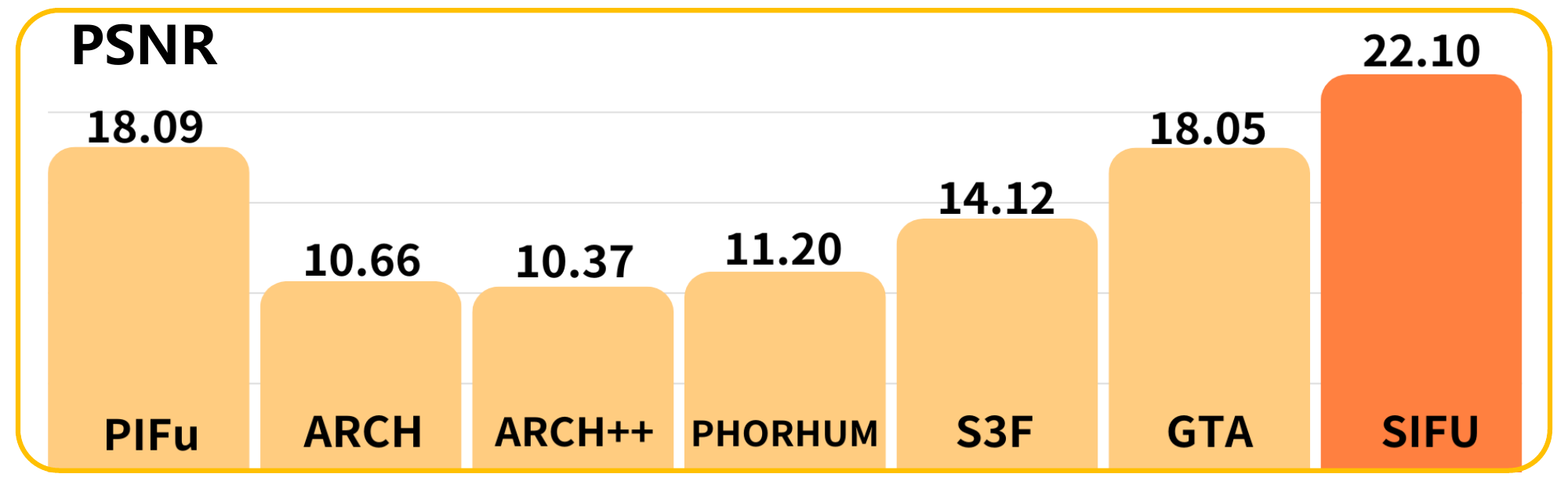}
        \caption{
            Quantitative comparison of texture quality on THuman2.0~\cite{THuman2.0:2021}.
        }
        \label{fig: psnr}
    \end{subfigure}
    \begin{subfigure}{\linewidth}
        \includegraphics[width=\linewidth]{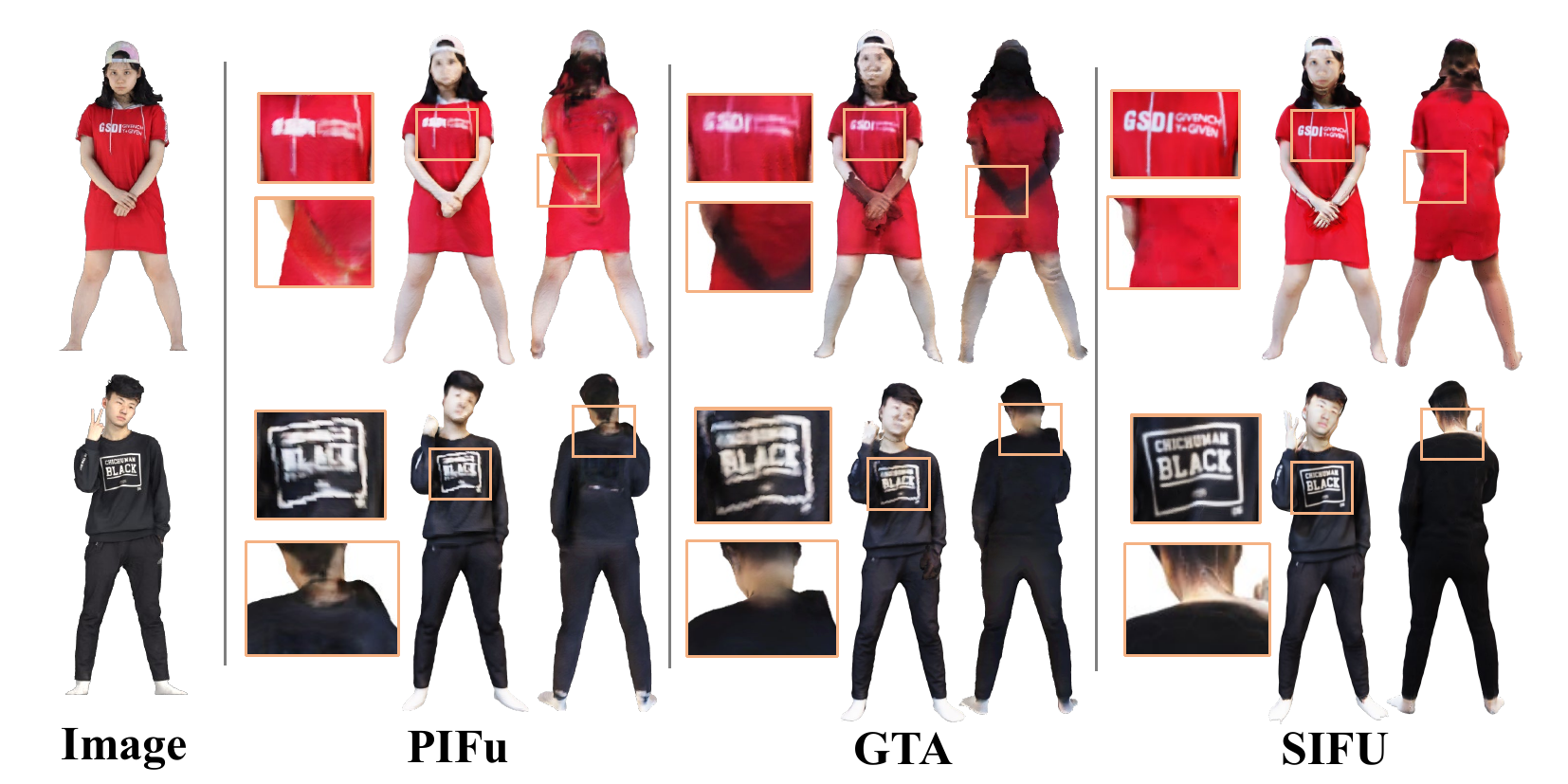}
	    \caption{
	   Qualitative results on Thuman2.0~\cite{THuman2.0:2021}
	    }
	    \label{fig: qualitative texture}
    \end{subfigure}}
    \vspace{-2.0em}
   \caption{\textbf{Texture comparison against SOTAs (\S\ref{sec:evaluation}).} We quantitatively and qualitatively compare texture quality on THuman2.0~\cite{THuman2.0:2021}. PIXIE~\cite{PIXIE:2021} used for SMPL-X estimation during testing. Please \faSearch~\textbf{zoom in} for details.}
    \vspace{-2.0mm}
\label{fig:tex_compare}
\end{figure}

\noindent\textbf{Quantitative Evaluation.} In geometry evaluation, our experiments utilize the ground-truth SMPL-X model for methods using a "SMPL-X body prior," as shown in \tab{table:geo-metrics}. \mn establishes a new standard in all metrics, especially excelling on the THuman2.0 dataset with an unprecedented Chamfer and P2S of \textbf{0.6 cm}. This highlights \mn's proficiency in accurate reconstructions across diverse scenarios, benefiting from our side-view conditioned approach.

For texture reconstruction, \mn surpasses PIFu~\cite{saito2019pifu} by \textbf{22.2\%} in PSNR, demonstrating its superior texture quality. For visual comparisons, refer to~\cref{fig:tex_compare} and the SupMat.

\noindent \textbf{Robustness to SMPL-X.} In real-world scenarios, encountering in-the-wild images lacking precise SMPL-X parameters is common. The ability to handle SMPL-X estimation errors is crucial for high-quality reconstructions. We evaluated our model's resilience by introducing noise (scaled by 0.05) to the pose and shape parameters of the ground-truth SMPL-X models. As shown in~\tab{table:ablation-smpl}, \mn demonstrates significant robustness, indicating strong practical utility.

\noindent \textbf{Qualitative Results.} Our results showcase the model's strong performance on in-the-wild images. As depicted in~\cref{fig: qualitative figure}, our model is capable of handling complex scenarios such as loose clothing and challenging poses with proficiency. Further examples are provided in the SupMat.

\subsection{Ablation Studies}\label{sec:ablation}

\noindent \textbf{Different Backbone Analysis.} In validating the effectiveness of our side-view decoupling transformer, we experimented with various alternative architectures. As per the results in~\tab{table:ablation}, self-attention and learnable embeddings, without SMPL-X guidance, led to significant errors, and even convolutional networks with similar capacities were unable to effectively link input images with SMPL-X conditioned views. This ablation study clearly demonstrates that our custom transformer architecture excels, delivering superior reconstruction results.

\begin{table}[t]
\centering
\scriptsize
\resizebox{\linewidth}{!}{
\begin{tabular}{cccc}
\shline
\rowcolor{gray!30} 
 \multicolumn{1}{c|}{Method} &
  Chamfer $\downarrow$ &
  P2S $\downarrow$ &
  \multicolumn{1}{c}{Normal $\downarrow$} \\ \shline
\multicolumn{4}{c}{\textit{A - Different Backbone}} \\ \hline
 \multicolumn{1}{c|}{no cross-attention} &  
  0.9846 &
  0.8672 &
  \multicolumn{1}{c}{0.0477}   \\
\multicolumn{1}{c|}{learnable embedding}&
   0.9860 &
  0.8538 &
  \multicolumn{1}{c}{0.0471} \\
\multicolumn{1}{c|}{use convolution network} & 
  0.8699 &
  0.8221 &
  \multicolumn{1}{c}{ 0.0387}\\
  \hline
\multicolumn{4}{c}{\textit{B - Different Feature Plane}} \\ \hline  
  \multicolumn{1}{c|}{only front plane} & 
  1.1165 &
  0.9574 &
  \multicolumn{1}{c}{0.0558}\\
  \multicolumn{1}{c|}{front and back planes} & 
  0.9929 &
  0.9189 &
  \multicolumn{1}{c}{0.0464}\\
  \multicolumn{1}{c|}{w/o left plane} & 
  \btwo 0.7941 &
  \btwo 0.7576 &
  \multicolumn{1}{c}{0.0387}\\
  \multicolumn{1}{c|}{w/o right plane} & 
  0.8058 &
  0.7671 &
  \multicolumn{1}{c}{\btwo 0.0386}\\
  \hline
\multicolumn{4}{c}{\textit{C - Different Query Strategy}} \\ \hline
\multicolumn{1}{c|}{pixel-aligned} &
   0.8111 &
   0.7615 &
  \multicolumn{1}{c}{0.0400} \\ 
  
\hline
\multicolumn{1}{c|}{\textbf{Ours}} &
  \bone \textbf{0.7725} &
  \bone \textbf{0.7354} &
  \multicolumn{1}{c}{\bone \textbf{0.0378}} \\ \shline
\end{tabular}}
\vspace{-1.0 em}
\caption{\textbf{Ablation study (\S\ref{sec:ablation}).}
We quantitatively evaluate the contribution of each component in our model. The evaluation is performed on the CAPE-NFP dataset, with ground-truth SMPL-X models provided during the testing phase.} 
\vspace{-1.0 em}
\label{table:ablation}
\end{table}

\begin{figure}[htbp]
\centering
\scriptsize
\includegraphics[width=\linewidth]{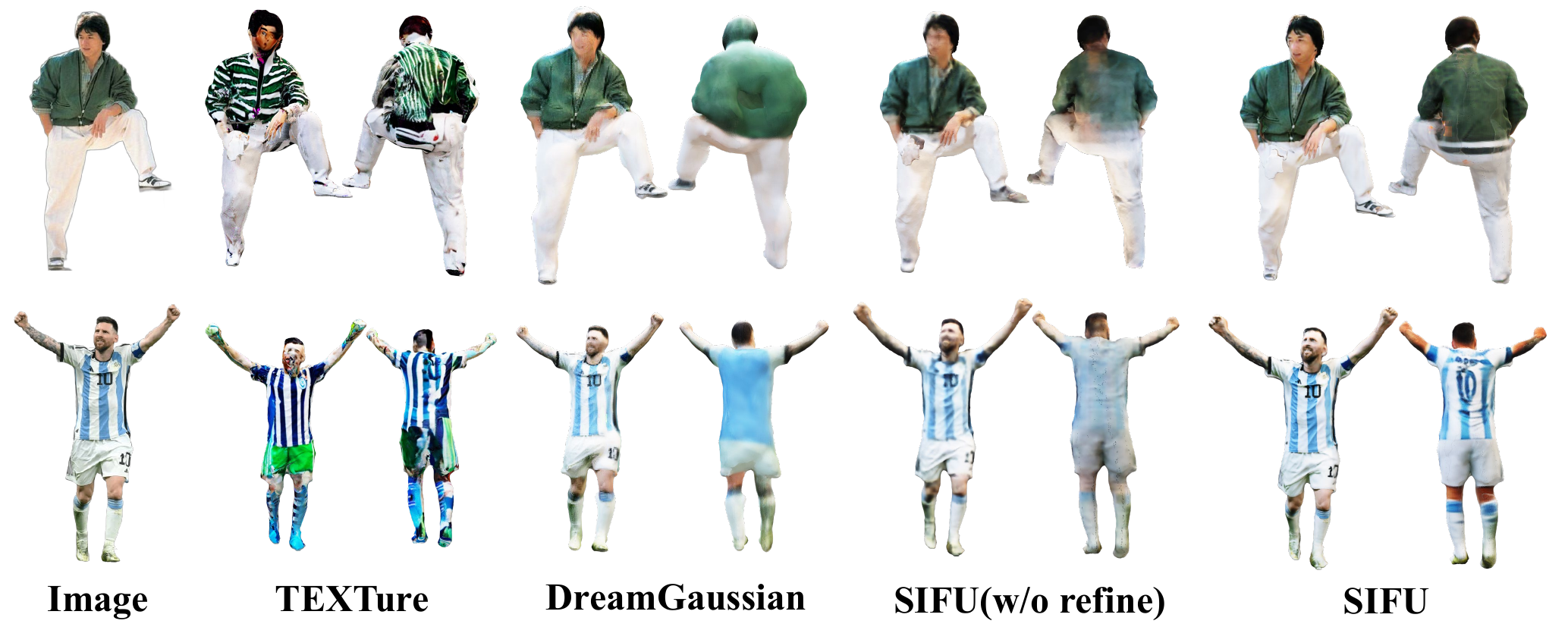}  
\vspace{-2.0 em}
\scriptsize
\caption{\textbf{Ablation on texture refinement (\S\ref{sec:ablation}).} We compare our 3D consistent texture refinement with other diffusion-based methods on in-the-wild images. Please \faSearch~\textbf{zoom in} to see details.}
\vspace{-0.5 em}
\label{fig:ablation_texture}
\end{figure} 

\noindent \textbf{Different Feature Plane Analysis.} In assessing the effect of various numbers of side-view feature planes, we found, as shown in~\tab{table:ablation}, that adding just the left or right side-view planes most significantly improved accuracy, reducing the Chamfer by about 0.2 cm. The inclusion of all four planes offered a smaller error reduction, approximately 0.03 cm. Considering the minor improvements from more planes against the added complexity, we chose a balanced approach with four planes, as shown in~\cref{fig:pipeline}.

\noindent \textbf{Query Strategy Efficacy.} We compared the hybrid prior fusion strategy with the pixel-aligned method~\cite{saito2019pifu,saito2020pifuhd,xiu2022icon}. As shown in~\tab{table:ablation}, the hybrid approach consistently outperforms the conventional method in all evaluation metrics.

\noindent \textbf{Different Texture Refinement.} In comparing our approach with diffusion-based methods like TEXTure~\cite{richardson2023texture} and DreamGaussian~\cite{tang2023dreamgaussian} (using Zero123 XL~\cite{liu2023zero1to3}), and also against our model without refinement, it is evident from~\cref{fig:ablation_texture} that our 3D Consistent Texture Refinement method excels in both texture quality and consistency.

\newcommand{\qualitativeCaption}{
\textbf{Qualitative results on in-the-Wild images (\S\ref{sec:evaluation}):} The first two rows present results for humans wearing loose clothing, and the subsequent two rows display outcomes for humans in challenging poses. (\faSearch~\textbf{Zoom in} for detailed view)
 }

\begin{figure*}[tbp]
    \centering
    \scriptsize
    \includegraphics[width=\linewidth]{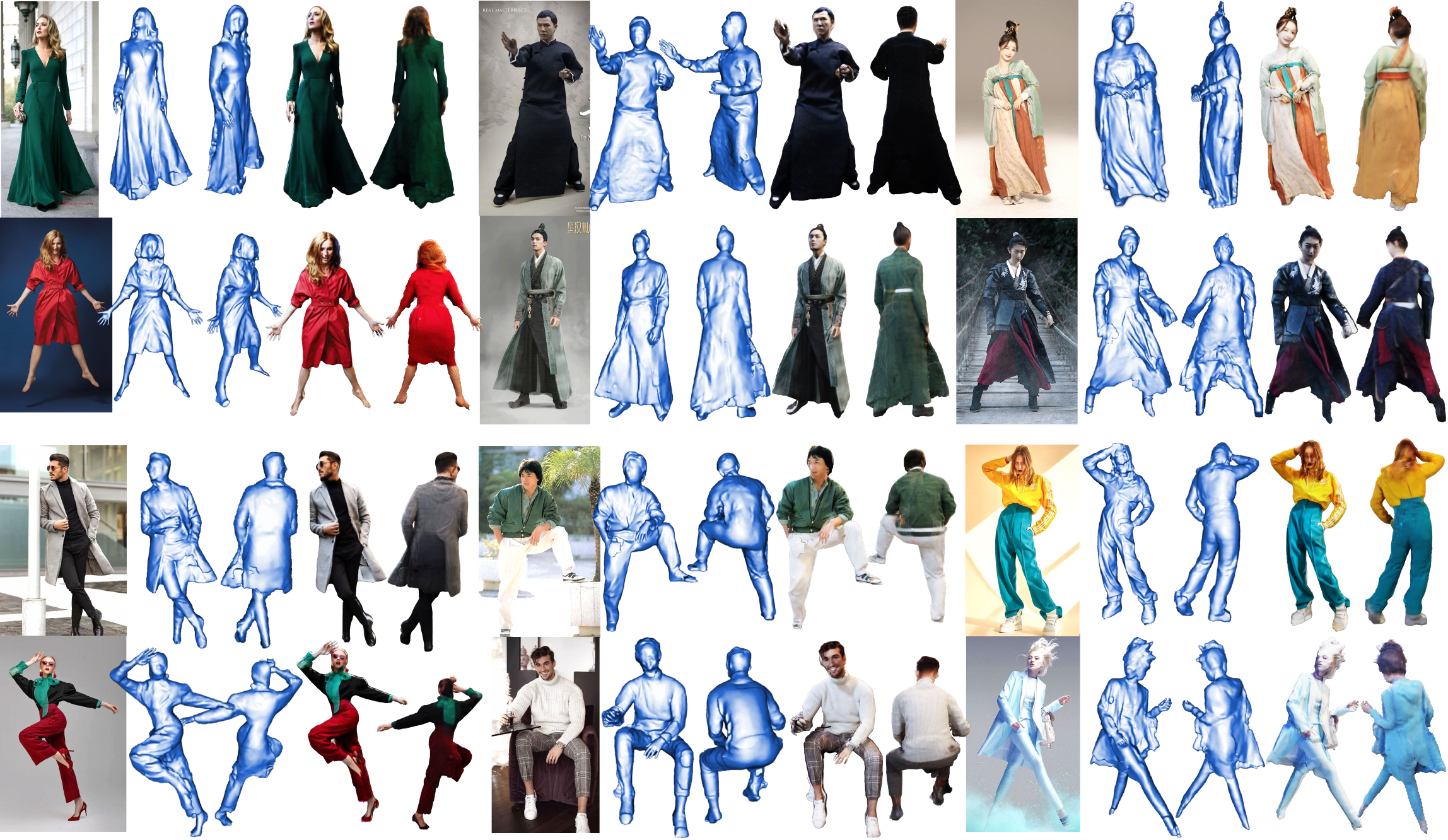}
    \vspace{-2.0em}
    \captionof{figure}{\qualitativeCaption}
    \label{fig: qualitative figure}
\end{figure*}

\subsection{Applications}\label{sec:application}

\begin{figure}[htbp]
\centering
\scriptsize
\includegraphics[width=\linewidth]{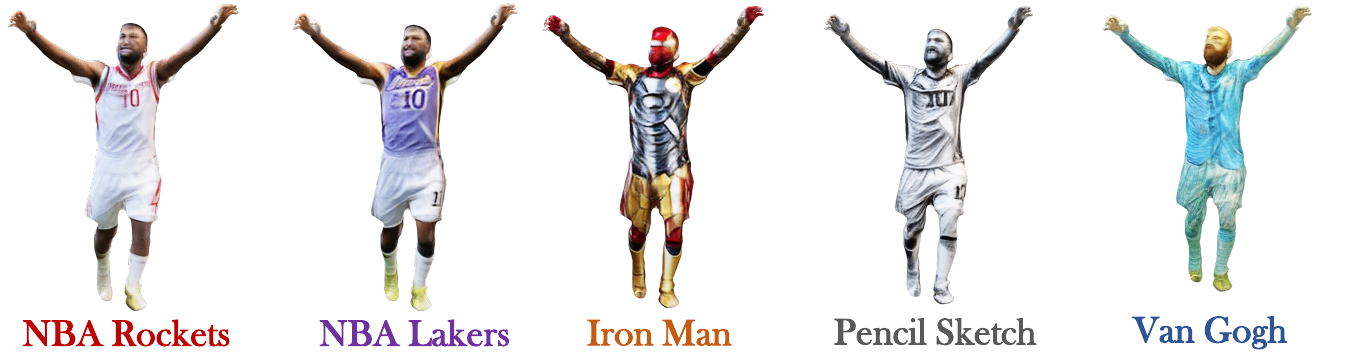}  
\vspace{-2.0 em}
\scriptsize
\caption{\textbf{Texture editing (\S\ref{sec:application}).} We edit the texture of the individual in~\cref{fig:pipeline} to achieve diverse outcomes by changing the text prompt in our 3D Consistent Texture Refinement.}
\vspace{-1.0 em}
\label{fig:tex_edit}
\end{figure}

\noindent \textbf{Texture Editing.} With the powerful ability of text-to-image diffusion models, we can change the text prompt to easily generate edited textures in the 3D consistent texture refinement. The edited results are shown in~\cref{fig:teaser} and~\cref{fig:tex_edit}.

\begin{figure}[htbp]
\centering
\scriptsize
\includegraphics[width=\linewidth]{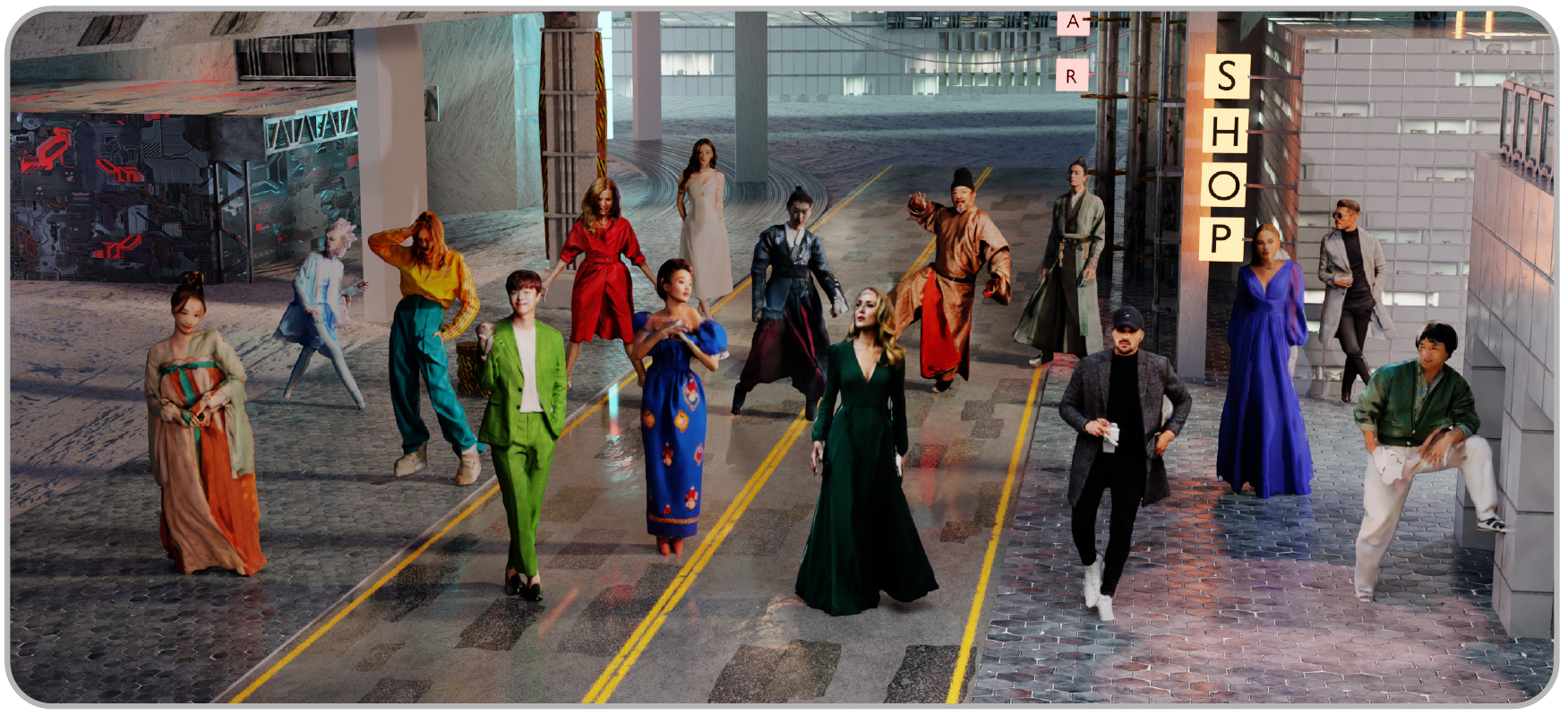}  
\vspace{-2.0 em}
\scriptsize
\caption{\textbf{Building scenes with SIFU reconstructed humans (\S\ref{sec:application}).} We showcase examples of building impressive scenes with SIFU reconstructed humans. Please \faSearch~\textbf{zoom in} to see details.}
\vspace{-2.0 em}
\label{fig:building scene}
\end{figure}

\noindent \textbf{Scene Building and 3D Printing.} The model's accurate geometry and refined textures make it ideal for virtual scene creation and 3D printing (see~\cref{fig:teaser,fig:building scene} and SupMat). It enhances realism in simulations and games and streamlines the 3D printing process, reducing the need for complex scanning. This has potential applications in rapid prototyping, educational resources, and custom 3D figurines.

\section{Conclusion}\label{sec:discussion_conclusion}

We introduce \mn, a novel method for reconstructing high-quality 3D clothed human meshes, complete with detailed textures. Our method employs SMPL-X normals~\cite{SMPL-X:2019} as queries in a cross-attention mechanism with image features, efficiently decoupling side-view features during the conversion of 2D features to 3D. This process significantly improves geometric accuracy and robustness in our 3D reconstructions. Moreover, we design a 3D Consistent Texture Refinement process, which employs text-to-image diffusion priors while maintaining consistency among diffusion features in the latent space. This innovative approach ensures the creation of realistic textures, particularly in regions that are not visible in the initial input. \mn distinctly outperforms existing methods in terms of both geometric and textural fidelity, showcasing exceptional capabilities in handling complex poses and loose clothing. These qualities make \mn highly suitable for real-world applications.

\noindent{\textbf{Acknowledgements.}} This work was supported by the National Natural Science Foundation of China (U2336212) and the Fundamental Research Funds for the Central Universities (No. 226-2022-00051).

\clearpage
{
    \small
    \bibliographystyle{ieeenat_fullname}
    \bibliography{main}
}
\maketitlesupplementary

\section{Implementation Details}
\label{sec:implementation}

\subsection{Model Structure}\label{sec:model structure}
\noindent\textbf{Side-view Conditioned Implicit Function.} We employ a Vision Transformer (ViT)\cite{ViT:2020} with a depth of 8 as our global encoder to transform the image into 2D latent features. To enhance the input, we concatenate the image with front and back normal maps, generated using an off-the-shelf model from\cite{xiu2022econ}. Our encoder, equipped with an 8-head multi-head self-attention mechanism, outputs features sized at $1024\times 256$.
For decoding, we utilize front and side decoders, each with a depth of 3. The front decoder applies multi-head self-attention to the encoded features, while the side decoders use multi-head cross-attention, taking SMPL-X normal images as queries. Both decoders operate with 8 attention heads. The SMPL-X normal images are converted into embeddings of $1024\times 256$ and are further enhanced with positional embeddings, ensuring effective feature extraction and attention. Ultimately, each decoder generates a feature map of dimensions $F\in \mathbb{R}^{128\times 128\times 32}$. 

In our comparative analysis, we evaluate the parameter efficiency of our feature extractor against established methods like PIFu~\cite{saito2019pifu} and GTA~\cite{zhang2023globalcorrelated}. As illustrated in~\tab{table:parameter}, our feature extractor uses fewer parameters compared to the transformer-based GTA method~\cite{zhang2023globalcorrelated}. This demonstrates the efficacy of our side-view decoupling transformer in extracting features more efficiently with a smaller parameter set.

\begin{table}[htbp]
\centering
\scriptsize
\resizebox{\linewidth}{!}{
\begin{tabular}{cccc}
\shline
\rowcolor{gray!30} 
 \multicolumn{1}{c|}{} &
 PIFu~\cite{saito2019pifu}&
  GTA~\cite{zhang2023globalcorrelated}&
  SIFU (ours) \\ \shline
\multicolumn{1}{r|}{Params}&
 24,744,136&
  45,060,544 &
  37,962,112  \\
  \shline
\end{tabular}}
 \vspace{-1.0 em}
\caption{\textbf{Parameters of the backbone network \S\ref{sec:model structure}}. }
\label{table:parameter}
\end{table}
\noindent In the Hybrid Prior Fusion Strategy, we use two identical MLPs for separate predictions of occupancy and color. Each MLP consists of layers with sizes [512, 1024, 512, 256, 128, 1]. Before inputting the integrated side-view features into the MLPs, we augment them with pixel-aligned normal features. These normal features are derived from off-the-shelf models referenced in~\cite{xiu2022econ} and are further refined using a 2-stack hourglass network~\cite{newell2016stacked}, resulting in $128\times 128\times 6$ normal feature dimensions.

\noindent\textbf{Image to Text Conversion.} We use image-to-text models to craft descriptive text prompts that facilitate the diffusion-based texture enhancement process. Specifically, we employ the advanced capabilities of GPT-4v~\cite{OpenAI_GPT4_2023} for generating precise descriptions of the input images. Upon uploading an image, we prompt the model with, "Please describe the person in the image in English, focusing on their clothing, colors, style, and hairstyle, without including the background. Limit the description to under 70 words." The generated description is represented as $A$. To tailor the prompts further, we prepend phrases like "the back side of $A$, realistic, vivid" to create a comprehensive input prompt $P$.

\noindent\textbf{Diffusion Model.} For our 3D Consistent Texture Refinement process, we employ the \textit{Stable Diffusion-v-2-1} checkpoint, accessed through HuggingFace. Throughout our experiments, we consistently use DDIM deterministic sampling, applying 50 steps. To invert the rendered views, we utilize DDIM inversion featuring a classifier-free guidance scale set to 1 and a sequence of 100 forward steps.

\begin{figure}[htbp]
\centering
\scriptsize
\includegraphics[width=.9\linewidth]{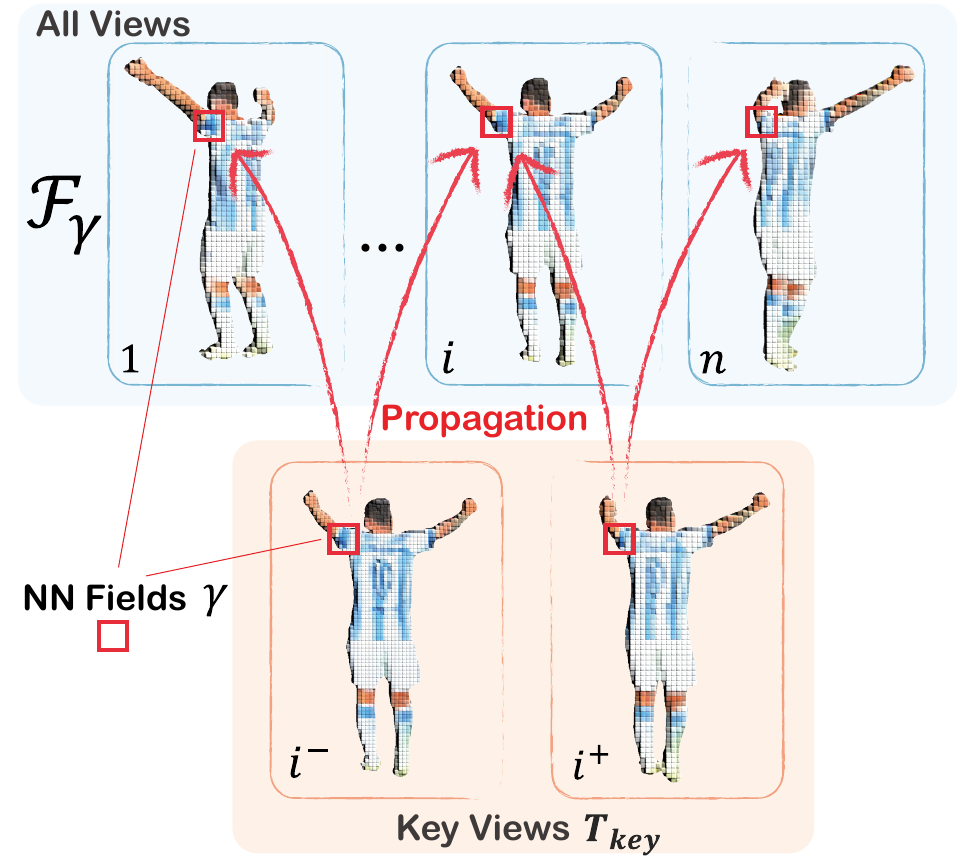}  
\scriptsize
\caption{\textbf{Consistent editing (\S\ref{sec:model structure}).}  
The modified features from key views are then propagated across all views based on the calculated nearest-neighbor (NN) fields.
}
\vspace{-1.0 em}
\label{fig:consistent editing}
\end{figure}

\noindent\textbf{Consistent Editing.}
Applying diffusion process directly to each rendered view may lead to content inconsistencies. Following~\cite{geyer2023tokenflow}, we attain consistent editing by enforcing consistency among the internal diffusion features across different rendered views. We carry out DDIM~\cite{song2020_ddim} inversion on $\mathcal{I}$ and extract the features $\phi(\boldsymbol{x}^i)$ from the self-attention module across every layer in the diffusion model. We then randomly select a set of key-views $\{J^i\}_{i\in \mathcal{\kappa}}$, and perform joint editing~\cite{wu2023tune} on these key views to obtain a set of features ${T}_{key}=\{\phi(J^i)\}_{i\in \mathcal{\kappa}}$ that exhibit a unified appearance. The features in ${T}_{key}$ will subsequently be propagated to all views.~\Cref{fig:consistent editing} briefly illustrates the entire process.


\newcommand{\SupsmplxCaption}{
\textbf{Qualitative comparison of robustness against SMPL-X estimation errors (\S\ref{sec:ablation details}).}
In this comparison, we evaluate the performance of various state-of-the-art methods in reconstructing figures under conditions of inaccurate SMPL-X estimation, which include challenges like bent legs and incorrect hand poses. Please \faSearch~\textbf{zoom in} for details.
 }

\begin{figure*}[tbp]
    \centering
    \scriptsize
    \includegraphics[width=\linewidth]{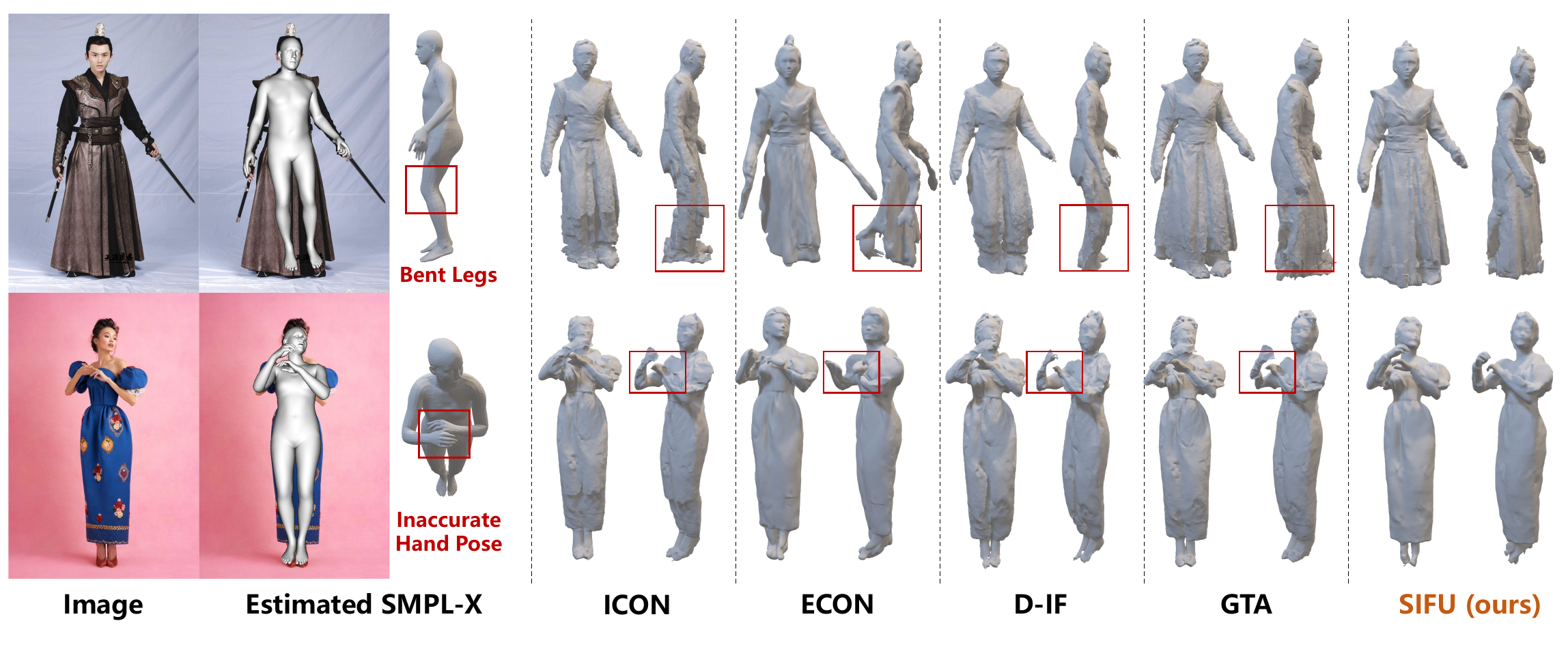}
    \vspace{-2.0em}
    \captionof{figure}{\SupsmplxCaption}
    \label{fig: sup smplx robust}
\end{figure*}

Upon processing the key views, we compute the nearest neighbors for the features of each view, ${\phi}(\boldsymbol{x}^{i})$, in relation to the features of its two adjacent key views, denoted as ${\phi}(\boldsymbol{x}^{i+})$ and ${\phi}(\boldsymbol{x}^{i-})$, where $i+$ and $i-$ are the indices of the closest next and past key-views, respectively. The resulting nearest neighbor fields are denoted as $\gamma^{i+}$, $\gamma^{i-}$:
\begin{equation}
\gamma^{i\pm}[p] = \argmin_{q}{\mathcal{D}\left({\phi(\boldsymbol{x}^i)[p]}, {\phi(\boldsymbol{x}^{i\pm})[q]}\right)}
\end{equation}
where $p,q$ represent spatial locations in the token feature map, and $\mathcal{D}$ signifies the cosine distance. For simplicity, we omit the notations of the generation timestep $t$ and the method is applied in all timesteps and self-attention layers. With $\gamma^{\pm}$ computed, we utilize it to propagate the edited features in ${T}_{key}$ to all other views. This propagation is executed by linearly blending the features in ${T}_{key}$ corresponding to each spatial location $p$ and view $i$:
\begin{equation}
\begin{split}
\mathcal{F}_{\gamma}({T}_{key},i, p) = \ & w_i \cdot \phi({J}^{i+}) [\gamma^{i+}[p]] \; + \\
& (1-w_i) \cdot \phi({J}^{i-}) [\gamma^{i-}[p]]
\end{split}
\end{equation}
where $\phi({J}^{i\pm}) \in {T}_{key}$ and $w_i \in (0,1)$ is a scalar proportional to the distance between view $i$ and its adjacent key-views, facilitating a smooth transition of features across views. Through this mechanism, we ensure consistency in the internal diffusion features across all rendered views during the editing process, significantly mitigating content inconsistencies and yielding a coherent editing outcome, especially in addressing invisible views.

\subsection{Data Processing and Training}\label{sec:training}
We generate training data using 3D scans from THuman2.0~\cite{THuman2.0:2021}. Each scan is rendered from 36 different angles at a resolution of 512, employing a weak perspective camera that horizontally rotates around the scan, under varying environmental lighting conditions. The model, implemented in PyTorch Lightning~\cite{Falcon_PyTorch_Lightning_2019}, is trained for 10 epochs with a learning rate of 1e-4 and a batch size of 4, over a span of 2 days on a single NVIDIA GeForce RTX 3090 GPU. During texture refining, we set $\lambda_1=1.0,\lambda_2=1e-4,\lambda_3=1e-2,\lambda_4=1.0$ for each loss item.

\noindent \textbf{Strategy for Point Sampling.} For each training subject, our approach involves obtaining 2048 points for occupancy, denoted as $G_o$, and 2048 points for color, symbolized as $G_c$. The method for occupancy point sampling is aligned with the strategy illustrated in ~\cite{saito2019pifu}. Color points are sampled uniformly, with a minor Gaussian disturbance, expressed as $\mathcal{N}(0, \sigma)$, wherein our experiment $\sigma$ is set at 0.1 cm. This disturbance occurs along the normals of the mesh surface. We obtain labels for the ground truth geometry, which specify whether a point is inside or outside the surface, through the application of Kaolin ~\cite{KaolinLibrary} to ascertain if a point lies within the ground truth mesh. The source of the ground truth color labels is the UV texture map of the 3D meshes.

\subsection{Inference}\label{sec:inference}
During the inference phase, we utilize Rembg~\cite{rembg} for background removal in in-the-wild images and employ PIXIE~\cite{PIXIE:2021} to estimate SMPL-X parameters, which are further refined following the process in~\cite{xiu2022icon}. The total time to reconstruct a clothed human mesh with coarse textures from an input image is approximately 44 seconds. In our 3D Consistent Texture Refinement step, we set the camera to render images in a weak perspective manner, with angles ranging from 150° to 210° in 2° increments, rendering 30 different viewpoints at 512 resolution to cover areas that are not visible. If finer textures are desired, the camera's range and resolution can be adjusted, though additional viewpoints will increase processing time. The texture refinement process takes about 5 minutes. As a result, we can generate a high-quality mesh with detailed textures in a total of about 6 minutes on a single NVIDIA GeForce RTX 3090 GPU.

During our tests on the CAPE~\cite{CAPE:2020} and THuman2.0~\cite{THuman2.0:2021} datasets, we utilize the ground-truth SMPL-X models, eliminating the need for parameter optimization. To assess reconstruction efficiency, we compared the time taken to generate a mesh from an input image on the CAPE-NFP dataset. As detailed in~\tab{table:time}, our model demonstrates time efficiency, being significantly faster than ECON~\cite{xiu2022econ} and on par with GTA~\cite{zhang2023globalcorrelated} in terms of running time.
\begin{table}[htbp]
\centering
\scriptsize
\resizebox{\linewidth}{!}{
\begin{tabular}{cccccc}
\shline
\rowcolor{gray!30} 
 \multicolumn{1}{c|}{} &
 ICON~\cite{xiu2022icon}&
 ECON~\cite{xiu2022econ}&
 D-IF~\cite{yang2023dif}&
  GTA~\cite{zhang2023globalcorrelated}&
  SIFU (ours) \\ \shline
\multicolumn{1}{r|}{Time (s)}&
  0.36&
  15.07 &
  0.40&
  0.68&
  0.65  \\
  \shline
\end{tabular}}
 \vspace{-1.0 em}
\caption{\textbf{Running time comparison on CAPE-NFP (\S\ref{sec:inference})}. }
\label{table:time}
\end{table}

\subsection{Ablation Details}\label{sec:ablation details}

\noindent \textbf{Robustness Against SMPL-X.} To simulate real-world robustness in our method, we introduce controlled random noise to both the $\boldsymbol{\beta}$ (body shape) and $\boldsymbol{\theta}$ (body pose) parameters of the SMPL-X model. We apply a noise scale of 0.05 to each, which is capable of inducing noticeable errors. For the shape parameters, $\boldsymbol{\beta}$, we generate random values corresponding to the number of $\boldsymbol{\beta}$ parameters, scale these from [0, 1) to [-1, 1) by subtracting 0.5 and then doubling, and subsequently apply the noise scale. This modified noise is then added to the original $\boldsymbol{\beta}$ values. We employ a similar process for the pose parameters, $\boldsymbol{\theta}$. The resulting imprecise SMPL-X models replace the ground truth for robustness evaluation in our approach.

As illustrated in~\cref{fig: sup smplx robust}, \mn demonstrates remarkable robustness in scenarios with inaccurate SMPL-X estimations in in-the-wild images, such as instances of bent legs and imprecise hand poses. Compared to state-of-the-art methods, \mn consistently produces satisfactory results even under these challenging conditions.

\noindent \textbf{Ablation Settings.} Our ablation study is structured into three distinct groups. In \textit{A - Different Backbone}, we explored the performance impact of replacing our custom transformer with various backbones. This included decoders based solely on self-attention, decoders that utilize learnable embeddings instead of SMPL-X queries (mirroring the approach in~\cite{zhang2023globalcorrelated}), and CNN-based hourglass networks~\cite{newell2016stacked} for processing across four views. In \textit{B - Different Feature Plane}, we assessed the efficacy of different feature planes by selectively disabling certain decoders. For instance, we used only the front decoder for scenarios labeled "only front plane" and a combination of front, back, and left decoders for the "w/o right plane" setup. Lastly, in \textit{C - Different Query Strategy}, we implemented a pixel-aligned strategy, projecting the query points onto the four feature planes, to evaluate the "pixel-aligned" approach.

\begin{figure}[htbp]
\centering
\scriptsize
\includegraphics[width=\linewidth]{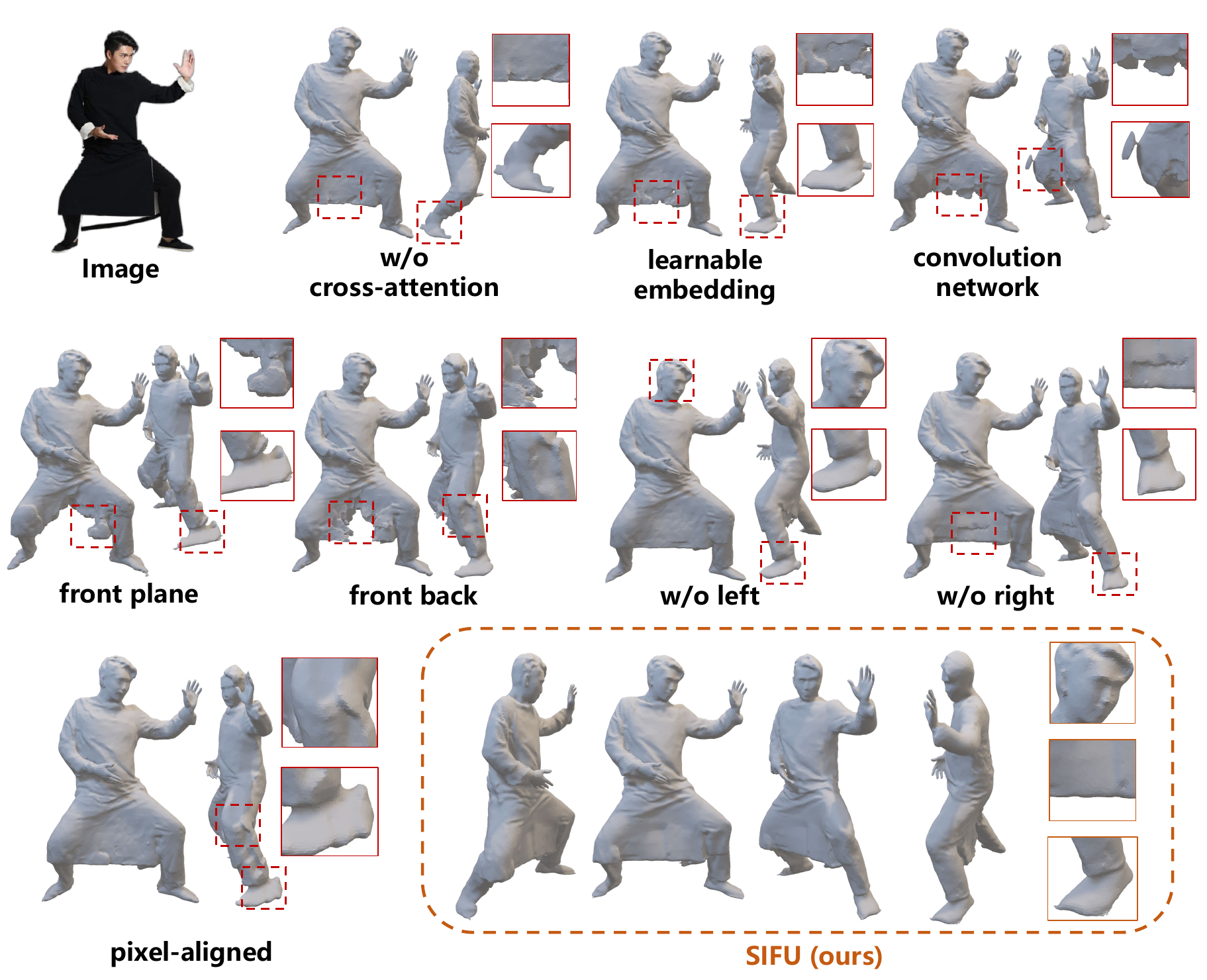}  
\scriptsize
\caption{\textbf{Qualitative comparison for ablation study (\S\ref{sec:ablation details}).} Please \faSearch~\textbf{zoom in} for details. }
\vspace{-1.0 em}
\label{fig:sup qualittive ablation}
\end{figure} 

In our qualitative analysis, we compared different ablation settings using the same input image. As depicted in~\cref{fig:sup qualittive ablation}, altering the network's backbone reveals noticeable artifacts, such as disrupted clothing and distorted human figures, leading to significant reconstruction errors. On the other hand, varying the feature planes demonstrates a clear progression in quality, starting from using only the front plane to incrementally including back and side planes. This improvement aligns with our quantitative findings. While using three feature planes (either excluding left or right) shows notable enhancements, there are still discernible discrepancies when compared to \mn, which leverages all four feature planes. Furthermore, the pixel-aligned method, lacking the SMPL-X prior constraint, tends to produce less realistic human-like parts.

\noindent \textbf{Quantitative Results on Texture Ablation.} In our quantitative assessment of the 3D Consistent Texture Refinement's impact, we employ 2D image quality metrics on multi-view colored images, rendered similarly to the normal images. These metrics include PSNR (Peak Signal-to-Noise Ratio), SSIM (Structural Similarity), and LPIPS (Learned Perceptual Image Patch Similarity). We also incorporate results from~\cite{albahar2023humansgd} for a more comprehensive comparison. As indicated in~\tab{table:texture table}, our refinement process enhances texture quality, particularly in terms of LPIPS, which more accurately aligns with human visual perception than traditional metrics like PSNR and SSIM. Moreover, our approach excels over baseline models in these metrics even before refinement. This performance can be attributed partly to our method's accurate color prediction and partly to the high-quality geometric reconstruction delivered by \mn.

\begin{table}[t]
\centering
\scriptsize
\resizebox{\linewidth}{!}{
\begin{tabular}{rcccc}
\shline
\rowcolor{gray!30} 
 \multicolumn{1}{c|}{Method} &\multicolumn{1}{c|}{Diffusion-based} &
  PSNR $\uparrow$ &
  SSIM $\uparrow$ &
  \multicolumn{1}{c}{LPIPS $\downarrow$} \\ \shline
\multicolumn{1}{r|}{PIFu~\cite{saito2019pifu}}&
\multicolumn{1}{c|}{\xmark}&
  18.0934 &
  0.9117 &
  \multicolumn{1}{c}{0.1372} \\
\multicolumn{1}{r|}{Impersonator++~\cite{liu2021liquid}} & 
\multicolumn{1}{c|}{\xmark}&
  16.4791 &
  0.9012 &
  \multicolumn{1}{c}{0.1468}\\
  \multicolumn{1}{r|}{TEXTure~\cite{richardson2023texture}} & 
  \multicolumn{1}{c|}{\cmark}&
  16.7869 &
  0.8740 &
  \multicolumn{1}{c}{0.1435}\\
\multicolumn{1}{r|}{Magic123~\cite{qian2023magic123}} &
\multicolumn{1}{c|}{\cmark}&
  14.5013 &
  0.8768 &
  \multicolumn{1}{c}{0.1880} \\ 
  
\multicolumn{1}{r|}{S3F~\cite{corona2023s3f}} & 
\multicolumn{1}{c|}{\xmark}&
   14.1212 &
  0.8840 &
  \multicolumn{1}{c}{ 0.1868}\\
  
\multicolumn{1}{r|}{HumanSGD~\cite{albahar2023humansgd}} & 
\multicolumn{1}{c|}{\cmark}&
   17.3651 &
   0.8946 &
  \multicolumn{1}{c}{ 0.1300}\\
  
  \hline
   
\multicolumn{1}{r|}{SIFU w/o refinement} &
\multicolumn{1}{c|}{\xmark}&
  \btwo 22.0256 &
  \btwo 0.9212 &
  \multicolumn{1}{c}{\btwo 0.0849} \\
  
\multicolumn{1}{r|}{SIFU} &
\multicolumn{1}{c|}{\cmark}&
  \bone \textbf{22.1024} &
  \bone \textbf{0.9236} &
  \multicolumn{1}{c}{\bone \textbf{0.0794}} \\
  
  \shline
\end{tabular}}
 \vspace{-1.0 em}
\caption{\textbf{Texture comparison on THuman2.0~\cite{THuman2.0:2021} (\S\ref{sec:ablation details}).}
During testing, PIXIE~\cite{PIXIE:2021} is used for SMPL-X estimation.
}
 \vspace{-2.0 em}
\label{table:texture table}
\end{table}

\section{Discussion}\label{sec:discussion}

\begin{figure}[htbp]
\centering
\scriptsize
\includegraphics[width=\linewidth]{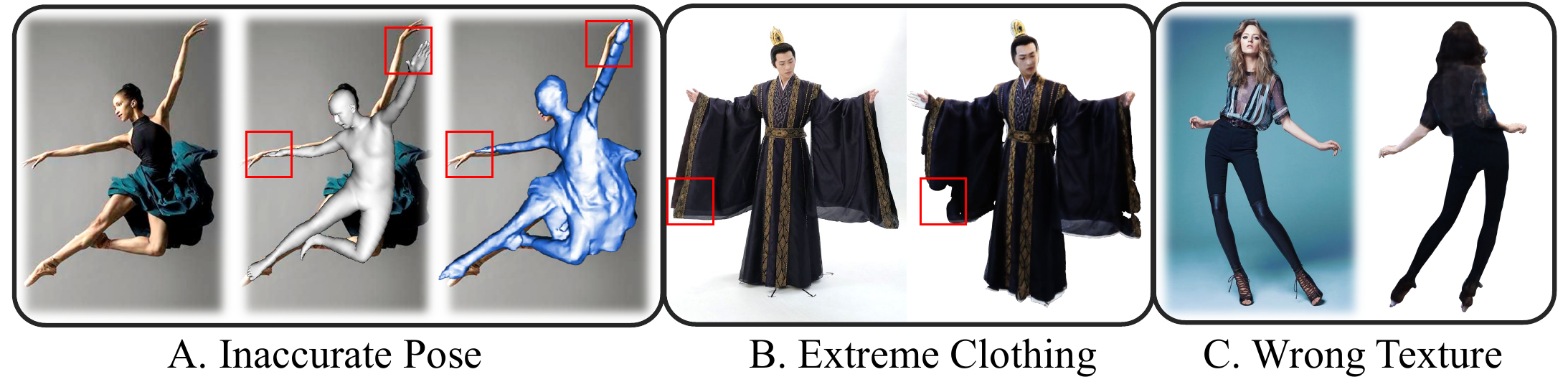}  
\scriptsize
\caption{\textbf{Failure cases of SIFU (\S\ref{sec:discussion}).} A. Inaccuracies in SMPL-X can compromise reconstruction precision. B. Fails with clothing significantly separated from the body. C. Text-driven editing can sometimes yield incorrect textures.}
\vspace{-1.0 em}
\label{fig:failure}
\end{figure}

\noindent\textbf{Limitations.}
While our model effectively reconstructs high-quality clothed human figures from single images, it encounters several challenges, as illustrated in~\cref{fig:failure}. Firstly, inaccuracies in SMPL-X estimation can compromise reconstruction accuracy, a limitation common in models reliant on human body priors. Secondly, our approach sometimes struggles with clothing that is distinctly separate from the body. Lastly, while translating images into text descriptions for the stable-diffusion model, some texture details may be lost, affecting the overall texture fidelity. This is because the nuanced complexities of an image might not be fully captured in text. To mitigate this, we utilize GPT-4v~\cite{OpenAI_GPT4_2023} for generating detailed and precise descriptions. Despite this, text-to-image diffusion models do not always perfectly replicate textures, presenting an area for potential improvement.

\noindent\textbf{Future Work}
In addition to overcoming current limitations, future research can explore several promising directions. Recent studies~\cite{liu2023zero1to3,lin2023magic3d,qian2023magic123,poole2022dreamfusion,tang2023dreamgaussian,huang2024tech} have demonstrated the potential of diffusion models in learning rich 3D priors, suggesting their integration in both shape and texture prediction. A significant challenge lies in effectively merging image-based visual prompts with the diffusion process while also aiming to reduce the time required for fine-tuning. Furthermore, employing distinct techniques for different parts of the human figure, such as hair, face, and hands, could yield more detailed results. The integration of these processed components could lead to the creation of highly detailed and comprehensive human models.

\noindent\textbf{Broader Impact.} 
Our model's ability to reconstruct 3D humans from single images raises concerns such as privacy infringement, potential for creating "deep fakes," and intellectual property violations. Addressing these issues necessitates collaborative efforts to establish ethical guidelines and legal frameworks, ensuring responsible use of this technology. It's vital to balance technological advancement with respect for individual rights and societal norms.

\section{Additional Results}\label{sec:additional results}

\begin{figure}[htbp]
\centering
\scriptsize
\includegraphics[width=\linewidth]{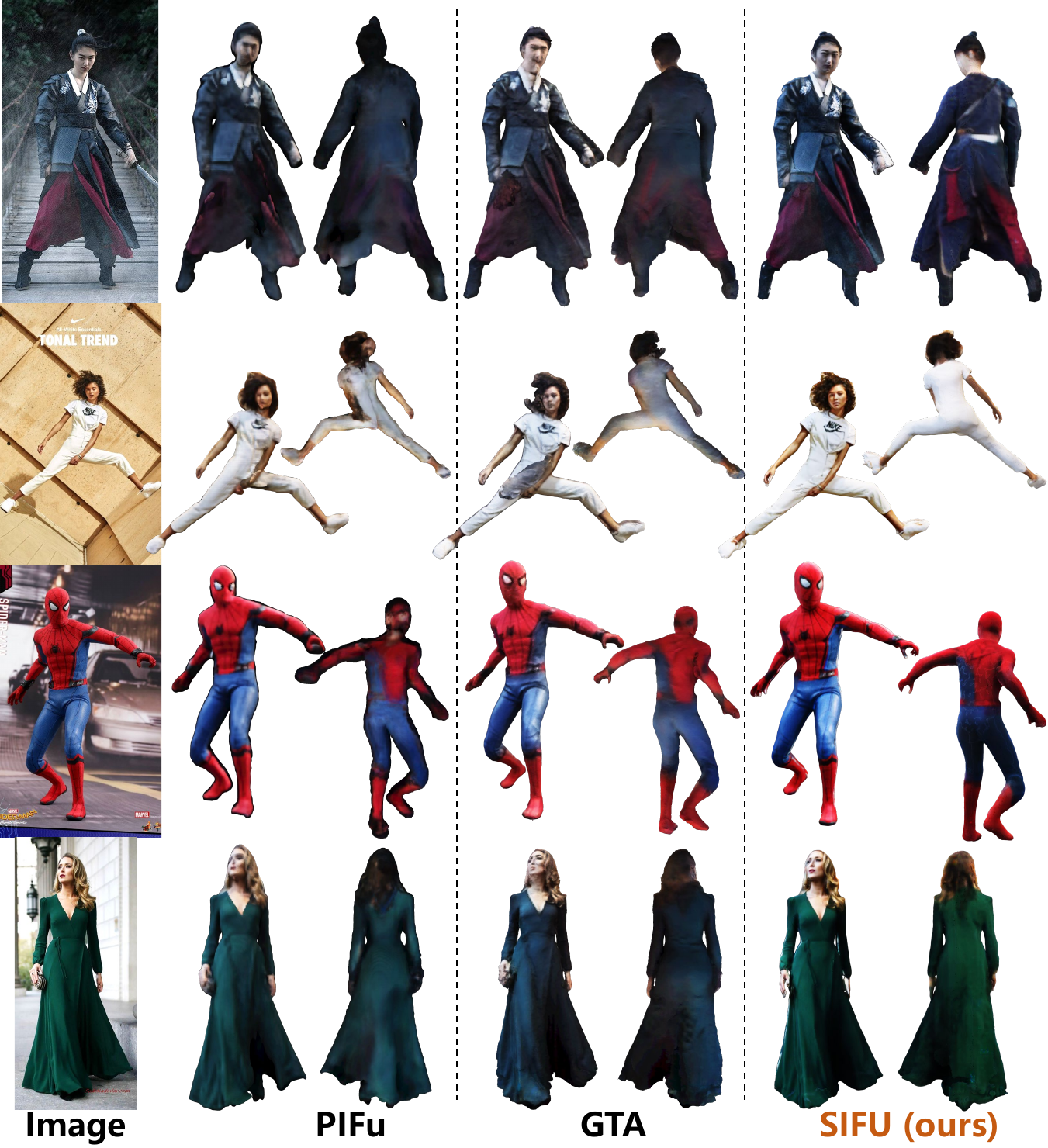}  
\scriptsize
\caption{\textbf{Qualitative comparison on texture quality (\S\ref{sec:additional results}).} }
\vspace{-1.0 em}
\label{fig:sup texture compare}
\end{figure}

\Cref{fig:sup texture compare} showcases a comparison of texture reconstruction results between \mn and current state-of-the-art methods. Previous approaches~\cite{saito2019pifu,zhang2023globalcorrelated} tend to produce suboptimal outcomes in invisible regions or exhibit blurry textures. In contrast, \mn effectively overcomes these issues, delivering consistently high-quality and visually coherent textures. 

\begin{figure*}[htbp]

\centering{
    \begin{subfigure}{\linewidth}
        \includegraphics[width=1.0\linewidth]{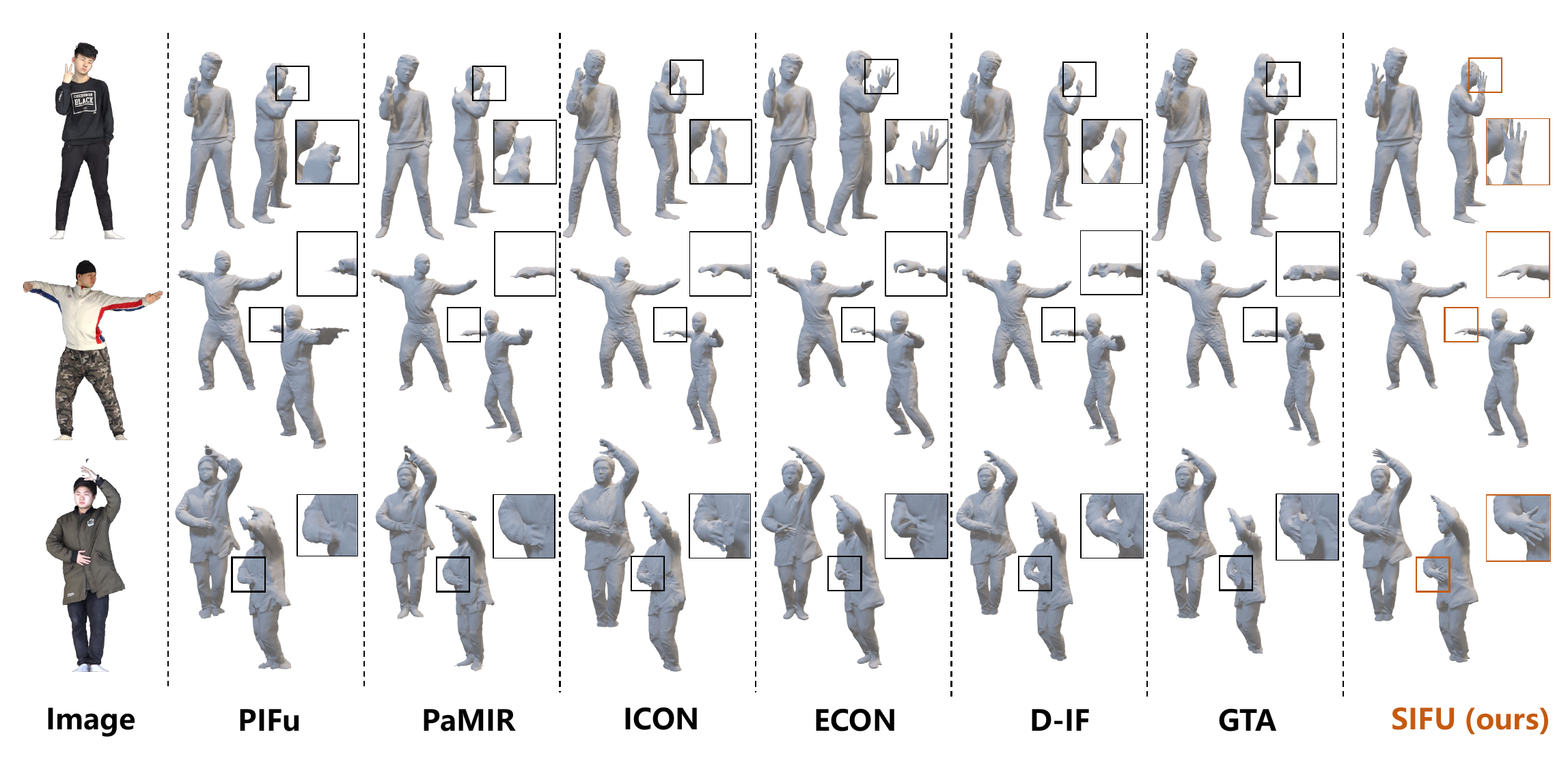}
        \caption{
            Qualitative comparison of geometry quality on THuman2.0~\cite{THuman2.0:2021}.
        }
        \label{fig: geometry thu}
    \end{subfigure}
    \begin{subfigure}{\linewidth}
        \includegraphics[width=\linewidth]{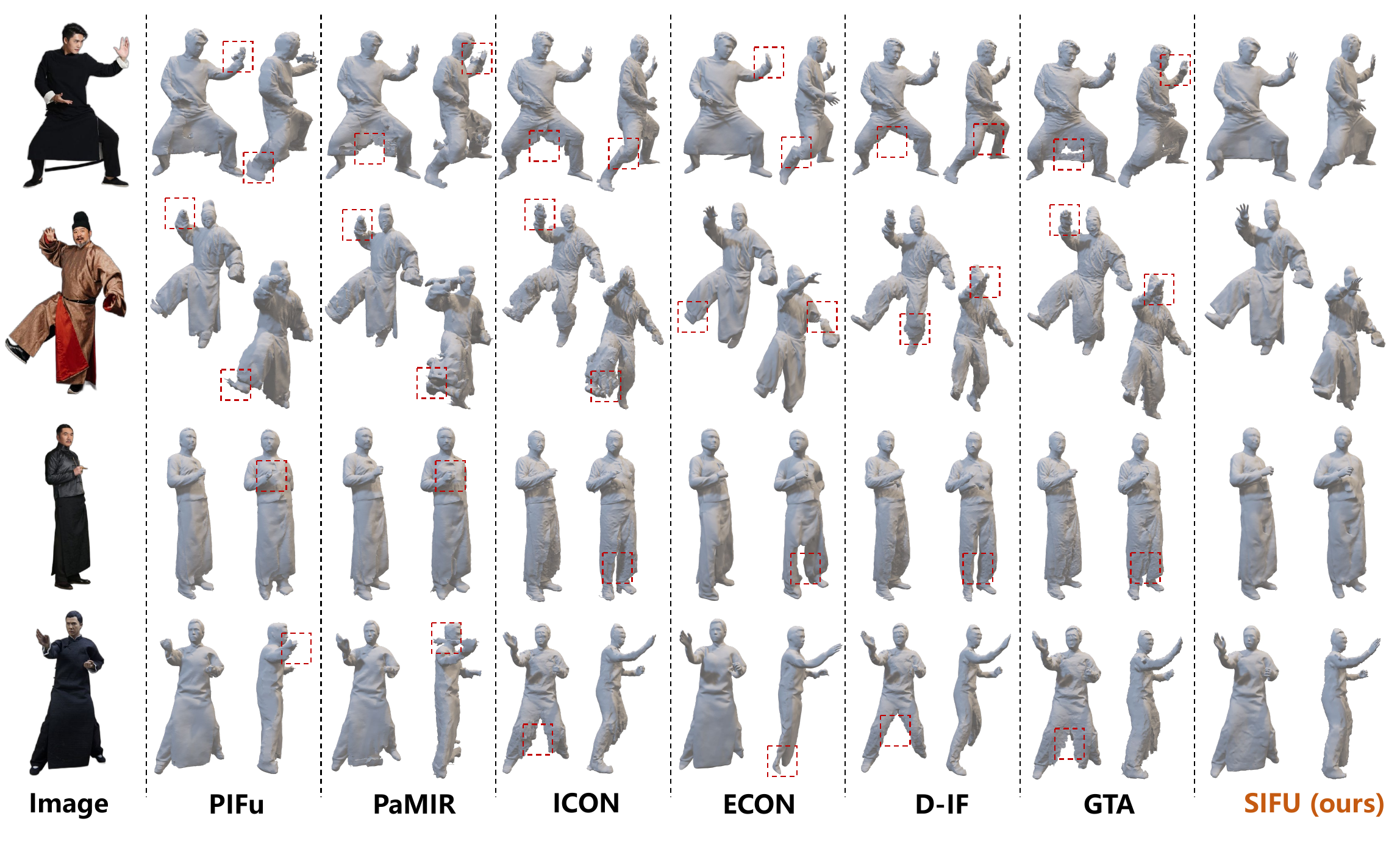}
	    \caption{
	   Qualitative comparison of geometry quality on in-the-wild images.
	    }
	    \label{fig: geometry wild}
    \end{subfigure}}
   \caption{\textbf{Geometry comparison against SOTAs.} Please \faSearch~\textbf{zoom in} for details.}
\label{fig:sup compare geometry}
\end{figure*}

Additionally, we assess the geometry reconstruction performance of \mn on both the THuman2.0 dataset~\cite{THuman2.0:2021} and in-the-wild images. As demonstrated in~\cref{fig:sup compare geometry}, \mn consistently delivers robust and high-quality reconstructions across a range of scenarios, including those with challenging poses and loose clothing. 
Finally, a diverse array of qualitative results on in-the-wild images is presented in~\cref{fig: sup qualitative 1,fig: sup qualitative 2}. These results encompass scenarios with complex poses, loose clothing, and vibrant textures, collectively demonstrating the extensive capabilities of \mn.


\clearpage

\newcommand{\SupqualitativeCaption}{
\textbf{Qualitative results on in-the-wild images (\S\ref{sec:additional results}).} 
 }

\begin{figure*}[tbp]
    \centering
    \scriptsize
    \includegraphics[width=\linewidth]{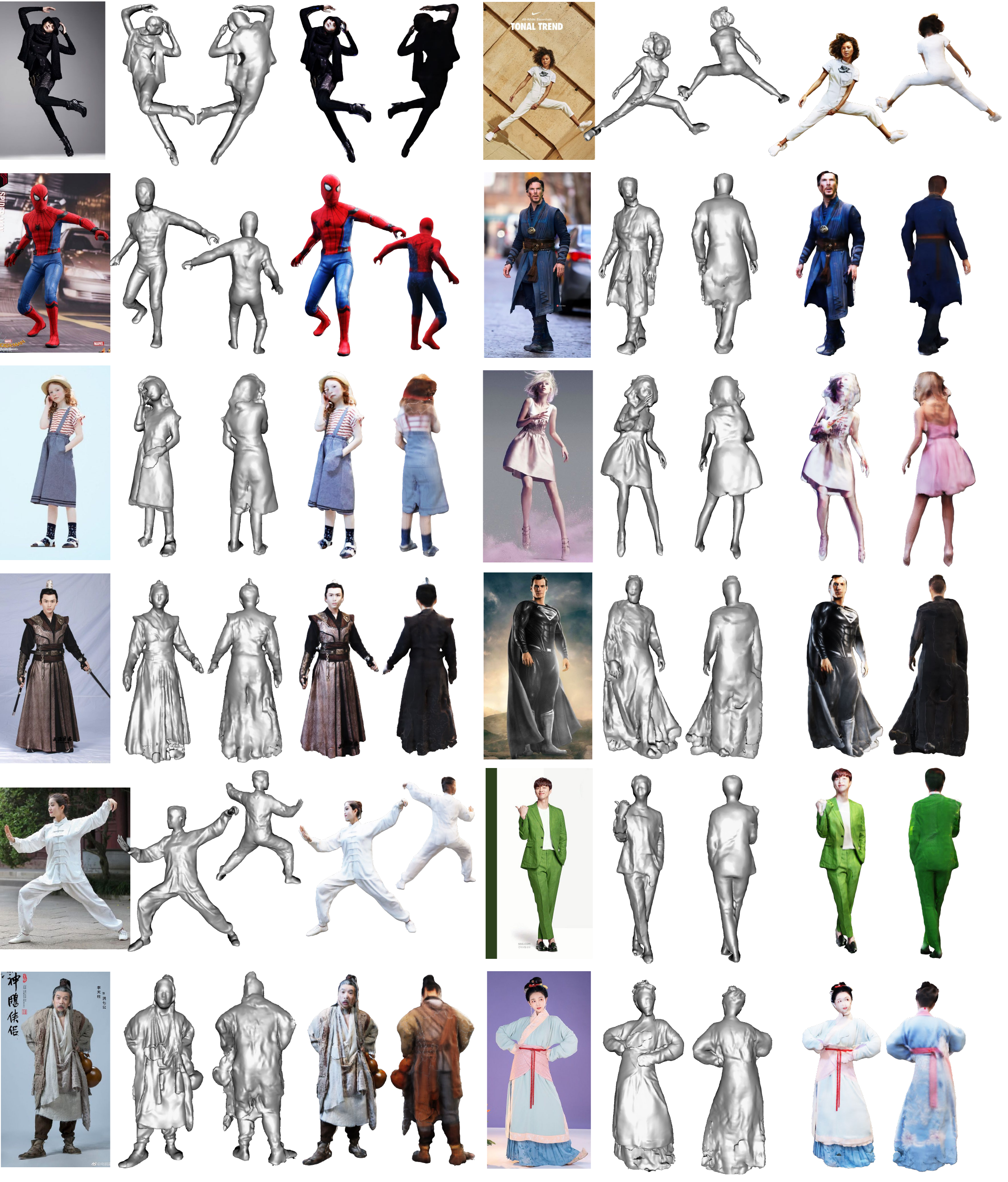}
    \vspace{-2.0em}
    \captionof{figure}{\SupqualitativeCaption}
    \label{fig: sup qualitative 1}
\end{figure*}

\newcommand{\Supqualitative}{
\textbf{Qualitative results on in-the-wild images (\S\ref{sec:additional results}).} 
 }

\begin{figure*}[tbp]
    \centering
    \scriptsize
    \includegraphics[width=.9\linewidth]{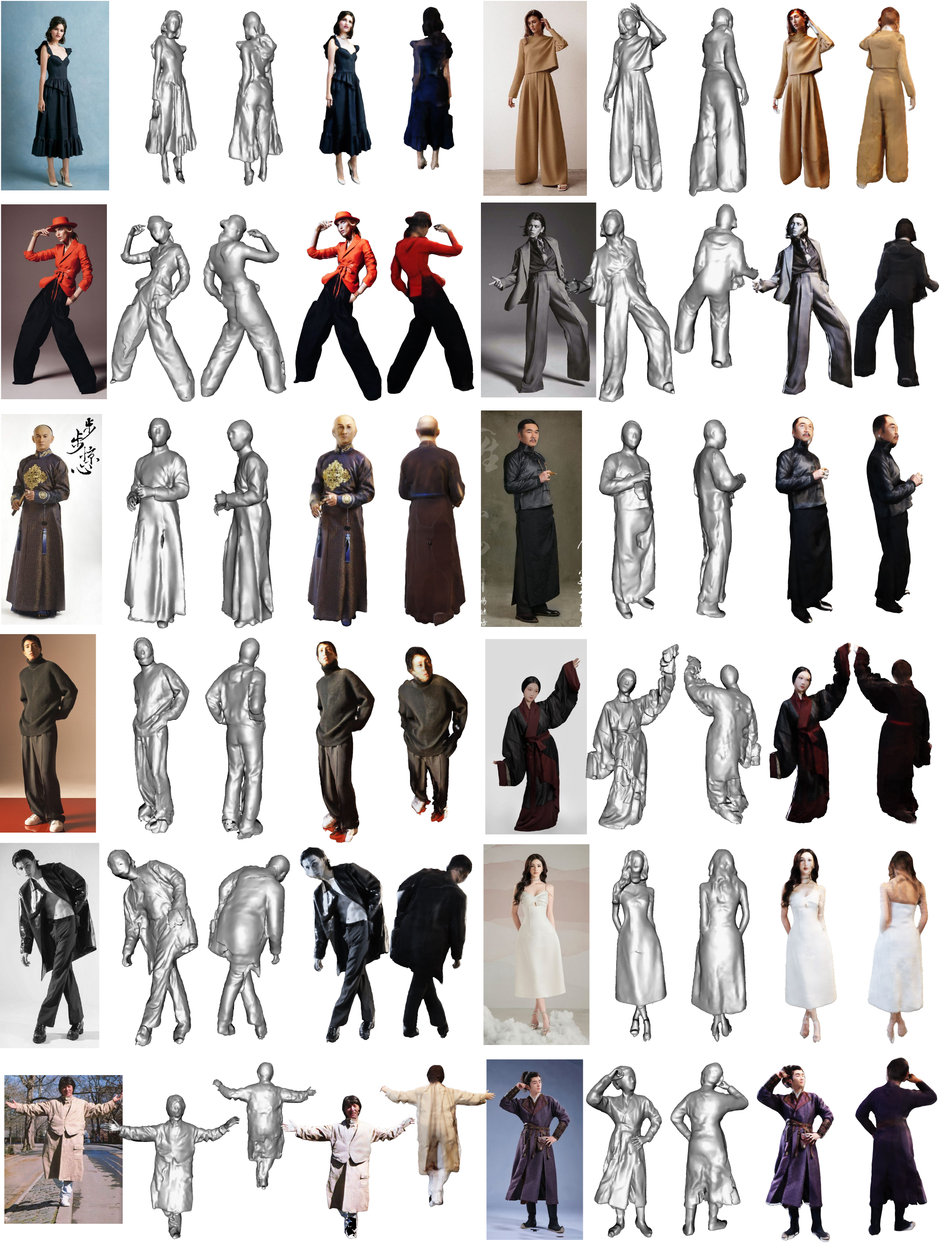}
    \vspace{1.0em}
    \captionof{figure}{\Supqualitative}
    \label{fig: sup qualitative 2}
\end{figure*}


\end{document}